\documentclass[conference]{IEEEtran}
\pdfoutput=1
\usepackage{times}
\usepackage{multirow}
\usepackage{sidecap}
\usepackage{afterpage}

\usepackage{color}
\usepackage{algorithm}
\usepackage{algorithmic}
\usepackage{amsmath}
\usepackage{amssymb}
\usepackage{wrapfig}
\usepackage{authblk}

\usepackage[numbers]{natbib}
\usepackage{multicol}
\usepackage[bookmarks=true]{hyperref}
\usepackage{graphicx}
\graphicspath{ {./figures/} }
\usepackage{titlesec}
\usepackage{subcaption}
\usepackage{multirow}
\usepackage{booktabs}
\usepackage{array}

\usepackage{soul}
\newcommand{\name}{\textit{SAR}}
\newcommand{\fullname}{\textit{Synergistic Action Representation}}
\newcommand{\datasetname}{\textit{Reorient} dataset}

\begin{document}

\title{SAR: Generalization of Physiological Agility and Dexterity via \textbf{S}ynergistic \textbf{A}ction \textbf{R}epresentation}

\author{Cameron Berg, Vittorio Caggiano$^*$, Vikash Kumar$^*$\\
\large Meta AI\\
\small{$^*$Equal advising}}

\makeatletter
\g@addto@macro\@maketitle{
  \begin{figure}[H]
  \setlength{\linewidth}{\textwidth}
  \setlength{\hsize}{\textwidth}
  \centering
  \resizebox{0.98\textwidth}{!}{\includegraphics[]{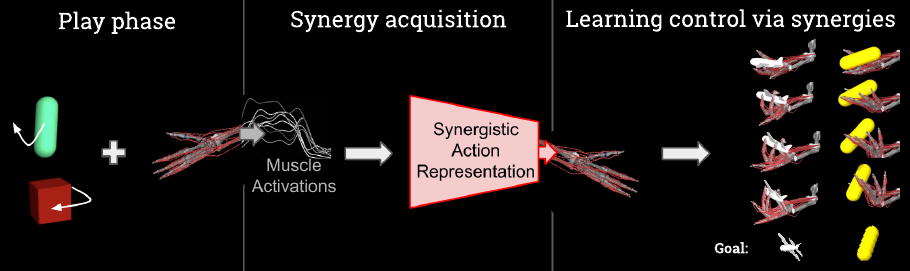}}
  \caption{\textbf{SAR pipeline.} A physiologically accurate musculoskeletal hand model is trained in simulation on a simple manipulation task (left), muscle activations are gathered from this policy (center), and these activations are used to create a synergistic action representation (SAR) that enables robust skill transfer for learning significantly more complex tasks (right). This pipeline is also used to train locomotor policies in a bipedal musculoskeletal model of human legs.}
  \label{fig1}
  \end{figure}
  \vspace{-7mm}
}
\maketitle{}
\begin{abstract}
Learning effective continuous control policies in high-dimensional systems, including musculoskeletal agents, remains a significant challenge. Over the course of biological evolution, organisms have developed robust mechanisms for overcoming this complexity to learn highly sophisticated strategies for motor control. What accounts for this robust behavioral flexibility? Modular control via muscle synergies, i.e. coordinated muscle co-contractions, is considered to be one putative mechanism that enables organisms to learn muscle control in a simplified and generalizable action space. 
Drawing inspiration from this evolved motor control strategy, we use physiologically accurate human hand and leg models as a testbed for determining the extent to which a \fullname~(\name{}) acquired from simpler tasks facilitates learning and generalization on more complex tasks. We find in both cases that \name{}-exploiting policies significantly outperform end-to-end reinforcement learning. Policies trained with \name{} were able to achieve robust locomotion on a diverse set of terrains (e.g., stairs, hills) with state-of-the-art sample efficiency (4M total steps), while baseline approaches failed to learn any meaningful behaviors. Additionally, policies trained with \name{} on in-hand 100-object manipulation task significantly outperformed ($\bold{>70\%}$ success) baseline approaches ($\bold{<20\%}$ success). Both \name{}-exploiting policies were also found to generalize zero-shot to out-of-domain environmental conditions, while policies that did not adopt \name{} failed to generalize. Finally, using a simulated robotic hand and humanoid agent, we establish the generality of SAR on broader high-dimensional control problems, solving tasks with greatly improved sample efficiency. To the best of our knowledge, this investigation is the first of its kind to present an end-to-end pipeline for discovering synergies and using this representation to learn high-dimensional continuous control across a wide diversity of tasks. 

\begin{center}
Project website: \url{https://sites.google.com/view/sar-rl}
\end{center}

\end{abstract}

\section{Introduction}

An important precondition for generalist embodied agents is their ability to exhibit diverse behaviors in response to dynamic environmental conditions, akin to humans and animals. From the practical perspective of constructing such agents, supporting behavioral diversity requires the integration of mechanisms that can capably handle complex high-dimensional action spaces. For example, humans are capable of an extremely large set of feasible motor repertoires, but this behavioral diversity necessitates continuous control of approximately 600 muscles in the human body \cite{gray_anatomy_1924}. 

Learning control policies for high-dimensional continuous action spaces are significantly more challenging \cite{rajeswaran_learning_2018} than learning in reduced action spaces that utilize idealized direct control, such as the Atari environments commonly used for reinforcement learning (RL) experiments \cite{kormushev_reinforcement_2013, bellemare_arcade_2013}. However, even the joint-based and end-effector continuous control dynamics commonly utilized in robotics exhibit significant simplifications as compared to the structural and functional physiology of actuated biological organisms \cite{salvietti_replicating_2018}. As a case study, consider the human hand, where joints are not controlled directly, but rather, through the proxy of pull-only muscle forces \cite{sobinov_neural_2021}. A further complication involved in physiological control is multiarticularity, meaning that there exist many-to-one and one-to-many relationships between muscles and joints (e.g., the flexor digitorum profundus muscle controls over 3 joints for each digit of the hand) \cite{kilbreath_distribution_2002}. Critically, the human physiology is also overactuated, meaning that there are more muscle forces than degrees of freedom (e.g., approximately 39 muscles collectively control 23 joints in the human hand) \cite{schumacher_dep-rl_2022}.

Despite the significant challenges inherent in learning musculoskeletal control, humans and animals are able to rapidly learn motor repertoires that are orders of magnitude more complex than the current state of the art in embodied agents\cite{balda_animal_1998}. A question naturally emerges from this fact: what strategies have humans and animals evolved that enable them to learn musculoskeletal control so robustly—and can these strategies be utilized to train policies that exhibit similarly robust behaviors in high-dimensional continuous action spaces?

Modular control is one mechanism known to be fundamental to human and animal behavioral flexibility, where motor output is computed as a product of a reduced number of modular spatiotemporal muscle coactivation patterns rather than as separate activations for each individual muscle \cite{caggiano_optogenetic_2016,bizzi_computations_1991}. These muscle co-contraction modules are referred to as \textit{muscle synergies}. There exists strong evidence not only that motor behavior is produced by coordinated muscle activity that can be represented with high fidelity in terms of a smaller number of synergistic activation patterns \cite{rabbi_non-negative_2020}, but also that the spinal cord explicitly encodes muscle activity at the level of synergies rather than as individual muscle activations \cite{bizzi_neural_2013}. This built-in action representation might be thought of as an embodied form of transfer learning, as it enables efficient adaptations of new motor repertoires that leverage similar—but non-identical—muscle activation patterns. It is suggested that these motor modules are present even in newborns \cite{dominici_locomotor_2011}, that they can be finetuned and recombined to yield more complex behaviors over the lifetime \cite{cheung_plasticity_2020}, and that they facilitate skill transfer, wherein similar activation representations can be reused for similar tasks (e.g., writing with a pen vs. writing with a pencil would not require learning two unique motor repertoires from scratch) \cite{bizzi_combining_2008}.

In this work, we explore whether the explicit discovery and utilization of muscle synergies improves the performance and generalization power of high-dimensional continuous control policies, with a specific focus on learning physiological control. In this investigation, we train physiologically accurate musculoskeletal models of the human hand (MyoHand) and legs (MyoLegs) \cite{caggiano_myosuite_2022} on a diverse task suite. To the best of our knowledge, this is the first study that demonstrates synergy-exploiting control policies that are able to successfully learn manipulation and locomotion tasks unsolved by baseline learning approaches. In particular, our core contributions are as follows:

\begin{itemize}
    \item We present a simple and generalizable paradigm to recover a synergistic action representation from behaviors learned during play.

    \item We demonstrate that training MyoLegs with a \fullname{} on a task suite of diverse terrains and conditions yields robust locomotion with state-of-the-art (SOTA) sample efficiency (3-5M total steps), while baseline approaches were found to learn no meaningful behaviors within this training budget.

    \item We further demonstrate that training MyoHand with a \fullname{} on a complex multi-object randomized manipulation task enables strong performance ($>70\%$ success) that is otherwise unsolved by baselines ($<20\%$ success).
    
    \item We show that both \name{}-exploiting policies generalize zero-shot to out-of-domain environmental conditions, while policies without access to this representation fail to generalize.

    \item Finally, we demonstrate that \fullname{}s extend beyond physiological control using robotic manipulation and full-body humanoid locomotion tasks, notably achieving SOTA performance with an order of magnitude improved sample efficiency on the latter.

\end{itemize}

\section{Related Works}
We first provide a background on what is currently understood about muscle synergies in humans and summarize previous work in leveraging synergy representations for learning musculoskeletal control.

\subsection{Motor neuroscience of muscle synergies}

Over five decades, the motor neuroscience literature on muscle synergies \cite{grillner_neurobiological_1985} has elucidated their mathematical representation \cite{tresch_matrix_2006, rabbi_non-negative_2020}, control dynamics \cite{saito_muscle_2018, ivanenko_five_2004}, lifespan finetuning and recombination \cite{dominici_locomotor_2011}, and cross-species similarity \cite{wainwright_evolution_2002, dominici_locomotor_2011}. Muscle synergies can be defined as motor activation modules that impose a dimensionality-reducing spatiotemporal structure to muscle activations by linearly combining (temporal) simplified activation patterns with a set of (spatial) variable activation weights for discrete muscles \cite{dominici_locomotor_2011}. It is widely believed that muscle synergies are an evolved mechanism for vertebrate motor control with a direct neurophysiological implementation \cite{song_neural_2022, hart_neural_2010, levine_identification_2014}. Accordingly, muscle synergies have been found to be robust to perturbations within a specific motor repertoire; for instance, as few as four muscle synergies in the lower limbs appear to control locomotion in humans and are invariant to movement speed, grade, direction, and skill \cite{dominici_locomotor_2011, saito_muscle_2018}. 

\subsection{Synergy representations for motor control}
The neuroscience literature on muscle synergies is reminiscent of dynamic movement primitives (DMPs), a popular concept in robotic control where actuation dynamics are decomposed into a smaller set of stable dynamical systems \cite{saveriano_dynamic_2021}. While these two concepts are similar in their simplification of complex continuous control dynamics into more tractable modular representations, the methods typically used to parameterize and stabilize DMPs can often preclude their generalization power to novel tasks \cite{ruckert_learned_2013}. By contrast, one core property of muscle synergies is their putative role as an action representation template for learning related tasks \cite{cheung_plasticity_2020}. 
 \begin{figure}
\includegraphics[width=.48\textwidth]{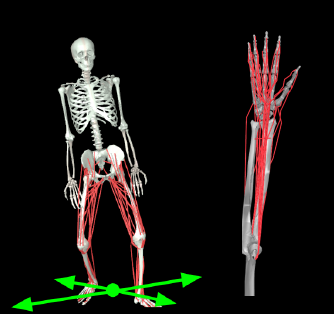}
  \caption{\textbf{MyoSuite musculoskeletal models}: \textit{MyoLegs} (16 DoF, 80 muscles) and \textit{MyoHand} (23 DoF, 39 muscles) \cite{caggiano_myosuite_2022}. Muscles are indicated in red.}
  \label{Fig:MyoHand}
\end{figure}

A small number of investigations have previously assessed the feasibility of utilizing muscle synergy representations for effectively learning to control musculoskeletal or robotic systems (for a review, see \cite{li_survey_2022}), albeit in fairly constrained task settings. To the best of our knowledge, all previous studies have explored whether synergy representations enable reaching behaviors in a musculoskeletal arm with 7 degrees of freedom (DoF) \cite{kutsuzawa_motor_2022, al_borno_effects_2020, diamond_reaching_2014, ruckert_learned_2013, chen_muscle-synergies-based_2021} or 8-DoF thumb/index model \cite{rombokas_reinforcement_2013}. These investigations have found that synergies can enable faster learning and partial generalization. One study also found that synergies improved task performance \cite{diamond_reaching_2014}. While these previous works demonstrate the general promise of leveraging muscle synergies for musculoskeletal control, the nature of these task demonstrations is arguably too low-dimensional both structurally (i.e., a 7-DoF arm) and functionally (i.e., reaching to a small number of fixed points) to yield strong conclusions about the full scope of using synergies for control.

In order to more comprehensively understand whether muscle synergies can enhance the training performance and generalization power of musculoskeletal control policies, we believe that synergy action representations must be tested in significantly more complex musculoskeletal task settings. For this reason, we investigate significantly beyond the current scope of the muscle synergy control literature by exploring (a) how synergistic action representations can be computationally extracted from simulated musculoskeletal models, and (b) how these representations can be leveraged to learn control policies capable of (i) achieving robust locomotion across a diverse set of terrains and (ii) manipulating thousands of complex objects. Finally, we investigate (c) the extent to which this biologically-inspired representation can more generally enable high-dimensional continuous control in task formulations beyond human physiology.
~~~
\section{Physiological Dexterity} 
\label{sec:problem_formualtion}

In this work, we seek to use learned action representations to acquire generalizable behavior across diverse conditions on physiologically accurate musculoskeletal models. We begin by introducing both models, \textit{MyoLegs} and \textit{MyoHand} and proceed to then formulate the problem of learning the associated behaviors for each model before detailing our experimental methods in Sec~\ref{sec:methods}. 
 
\subsection{Physiologically accurate musculoskeletal models}

We investigated various options of musculoskeletal models \cite{McFarland_2022, Lee_msk_2015} before ultimately selecting the \textit{MyoLegs} and \textit{MyoHand} models in favor of their computational efficiency, physiological accuracy, and support for contact dynamics, which is essential for the stated goal of yielding generalizable manipulation and locomotion behaviors \cite{caggiano_myosuite_2022}. Both models are implemented in the MuJoCo physics simulator \cite{todorov_mujoco_2012} and also exhibit important mechanistic features that differentiate biological actuation from the joint-based control approach traditionally utilized in robotics, including overactuation, multi-articulation, and pull-only actuation with third order muscle dynamics \cite{sobinov_neural_2021}. \textit{MyoLegs} is a physiologically realistic model of adult left and right legs (see Fig. \ref{Fig:MyoHand}), comprising 20 degrees of freedom and and 80 muscle-tendon units \cite{caggiano_myosuite_2022}. \textit{MyoHand} is a physiologically realistic model of an adult right hand-wrist (see Fig. \ref{Fig:MyoHand}), comprising 23 joints, 29 bones, and 39 muscle-tendon units \cite{caggiano_myosuite_2022}. Relevant to the current investigation, previous work has also demonstrated that \textit{MyoHand} can learn dexterous manipulation behaviors with single objects (e.g., twirling a pen to a desired orientation) \cite{caggiano_myosuite_2022}. Here, we aim to go beyond isolated task-specific single-object solutions, by using the \textit{MyoHand} model to learn in-hand manipulation behaviors that generalize from prior experiences to handle new, complex tasks.
 
Before outlining our methods, we next turn to formalizing our problem formulation. 

\subsection{Problem formulation}
We pose the problem of learning with \textit{MyoLegs} and \textit{MyoHand} as Markov Decision Processes (MDP) \cite{sutton_reinforcement_2018}, which can be defined as $ \mathcal{M} \ = (\mathcal{S},\mathcal{A},\mathcal{T},\mathcal{R},\mathcal{\rho},\mathcal{\gamma})$, where $\mathcal{S} \subseteq \mathbb{R}^{n}$ and $\mathcal{A} \subseteq \mathbb{R}^{m}$ each represent continuous state and action spaces. $\mathcal{T}$ represents the unknown distribution that describes state transition dynamics such that $s^{\prime} \sim \mathcal{T}(\cdot \mid s, a)$. The reward function is denoted as $\mathcal{R}: \mathcal{S} \rightarrow \left[0, R_{\max }\right]$, $\gamma \in[0,1)$ indicates the reward discount factor, and $\rho$ denotes the initial distribution of states. 
The goal in performing RL is to search for policy parameters $\theta$ that map from states to a probability distribution over actions $\pi_{\theta}: \mathcal{S} \rightarrow P(\mathcal{A})$ in order to maximize long-run discounted returns \cite{sutton_reinforcement_2018}. We optimize our policy using Soft Actor-Critic (SAC) \cite{haarnoja_soft_2018}, an off-policy RL algorithm used for continuous control which optimizes for both long-run discounted returns, $\sum_{t} \gamma^t R_t$, and policy entropy, $H(\pi)$, such that $\pi_\theta^*(a \mid s)=\operatorname{argmax}_{\theta}[J(\pi, \mathcal{M})]$, where

\begin{equation*}
J = \max_{\theta} \mathbb{E}\left[\sum_{t} \gamma^t\left(R\left(s_t, a_t, s_{t+1}\right)+\alpha H\left(\pi\left(\cdot \mid s_t\right)\right)\right)\right].
\end{equation*}

We later demonstrate (Fig. \ref{fig:reorient100_main}) that the direct optimization of the above objective in an end-to-end learning paradigm does not lead to meaningfully generalizable behavior. We attribute this failure to the challenges in navigating an extremely high-dimensional search space that hosts all internal degrees of freedom of the hand and the object under manipulation in addition to third-order actuation dynamics, intermittent contact dynamics, and the high likelihood of catastrophic failures due to uncoordinated movements during early exploration. 

In contrast to end-to-end learning, biological organisms faced with similar search spaces are easily able to synthesize sophisticated movements that are robust to this complexity. In the next section, we outline our method that takes inspiration from mechanisms that putatively contribute to this evolved behavioral sophistication.

\section{\name: Synergistic Action Representation} 
\label{sec:methods}

Exploration challenges in high-dimensional action manifolds naturally associated with manipulation and locomotion are well-recognized \cite{bicchi2000robotic}, including in the field of robotics \cite{bicchi2000robotic,rajeswaran_learning_2018, chen_system_2021} and computer vision \cite{zhang2021manipnet, mordatch2012contact}. 
Multiple techniques leveraging expert demonstrations \cite{rajeswaran_learning_2018}, curriculum learning \cite{openai_solving_2019, chen_system_2021}, and even human-designed mappings to embed representations into the learning process \cite{kumar_real-time_2014} have been proposed in order to aid and accelerate the acquisition of competent behavior. In contrast to these methods, which lean heavily on experts, we investigate if there exist learning paradigms that do not require such interventions at any point. Our work is heavily inspired by the increasingly strong evidence that the biological motor system enables actuation by leveraging synergistic control of the musculoskeletal system \cite{caggiano_optogenetic_2016, bizzi_combining_2008, davella_shared_2005, overduin_microstimulation_2012}. We proceed to ask:
\begin{enumerate}
    \item How can synergistic representations be automatically acquired?
    \item How can synergistic control can be embedded in behavior acquisition paradigms?
\end{enumerate}

\subsection{Synergy Acquisition}
\label{syn_ac}
Researchers in the fields of biomechanics and neuroscience have long pursued the search for synergies both at the joint-postural and muscle-activation levels. Due to the relative ease of recording joint posture over muscle activations, synergies at the level of joint postures \cite{todorov2004analysis, santello_postural_1998, della_santina_postural_2017} have been more extensively used to synthesize behaviours in robotic control \cite{ficuciello_synergy-based_2016,li_survey_2022}.  Nevertheless, although muscle synergies provide direct insight into biological motor control and learning \cite{davella_combinations_2003, davella_shared_2005, chen_muscle-synergies-based_2021, cheung_plasticity_2020}, they can only be practically extracted from a limited number of muscles in human legs and hands. This practical limitation represents the biggest limitation to directly using muscle synergies to synthesize behaviors \cite{taborri_feasibility_2018}.

How can we address this challenge to compute and extract synergies for the \textit{MyoLegs} and \textit{MyoHand} at the actuation level? We consider the underlying properties of synergies during locomotion and manipulation: synergies act like lower-dimensional submanifolds in the high-dimensional space that are consistently visited during the course of the behavior. In order to capture these recurrent pathways, we subject \textit{MyoLegs} and \textit{MyoHand} to a play phase, during which relatively simple behaviors are learned.

After the play period, we record muscle activation time series data from the resulting policies, $M^{act} \in \mathcal{R}^{a \times t}$. We posit that owing to the diversity of the behaviors captured in the dataset, its visitation density has a high degree of overlap with the synergistic submanifolds. While there are multiple plausible strategies for recovering submanifolds from such datasets, we resort back to the techniques common in motor neuroscience investigations of synergies from experimental datasets \cite{rabbi_non-negative_2020}. Applying both PCA and ICA to the time series data (taken together, ICAPCA) is accepted as a robust method for capturing muscle synergies \cite{tresch_matrix_2006}. In this work, we take the additional step of normalizing this output in the range [-1, 1] in order to reconfigure the resultant submanifold as a control signal.

$$SAR := |T_{ICA}(T_{PCA}(M^{act}))|$$ 

The resulting linear factorization is what we refer to as the \fullname{} (\name). Practically, this representation can also be viewed as a way to inverse transform any point in lower-dimensional synergy space to full muscle activation space $a_{full}^{syn} = SAR(a_syn)$. While $syn=full$ constitutes a complete reconstruction of the original signal, lower $syn$ will allow for capturing the most informative muscle co-activation patterns over time. For both \textit{MyoLegs} and \textit{MyoHand}, we define \name{} based on $syn = 20$ (which for both models captured over 80\% of the variance in $M^{act}$, see Appendix: Fig \ref{fig:synergies_VAF}).

\subsection{Behavior Acquisition with \name}\label{sec:behav_acquisition}

While \name{} encodes preferred muscle synergies, we still require a method for training locomotion and manipulation behaviors. The question thus becomes: how can use embed \name{} into a behavioral learning paradigm? One intuitive strategy is to learn a policy directly in the space of synergies and project it to full muscle activation space via \name{}:  $a^{syn}_{39} = SAR(\pi(a_N|\o_t))$. While the reduced dimensionality helps with the exploration challenges, it also theoretically restricts the policy from expressing behaviors significantly outside of the synergistic manifolds. Accordingly, we develop a policy architecture that preserves the benefits of directed search from the synergistic pathways but is still able to optionally learn task-specific behavioral finetuning in the original nonsynergistic manifold.

\begin{figure*}
\centering
\includegraphics[scale=.7]{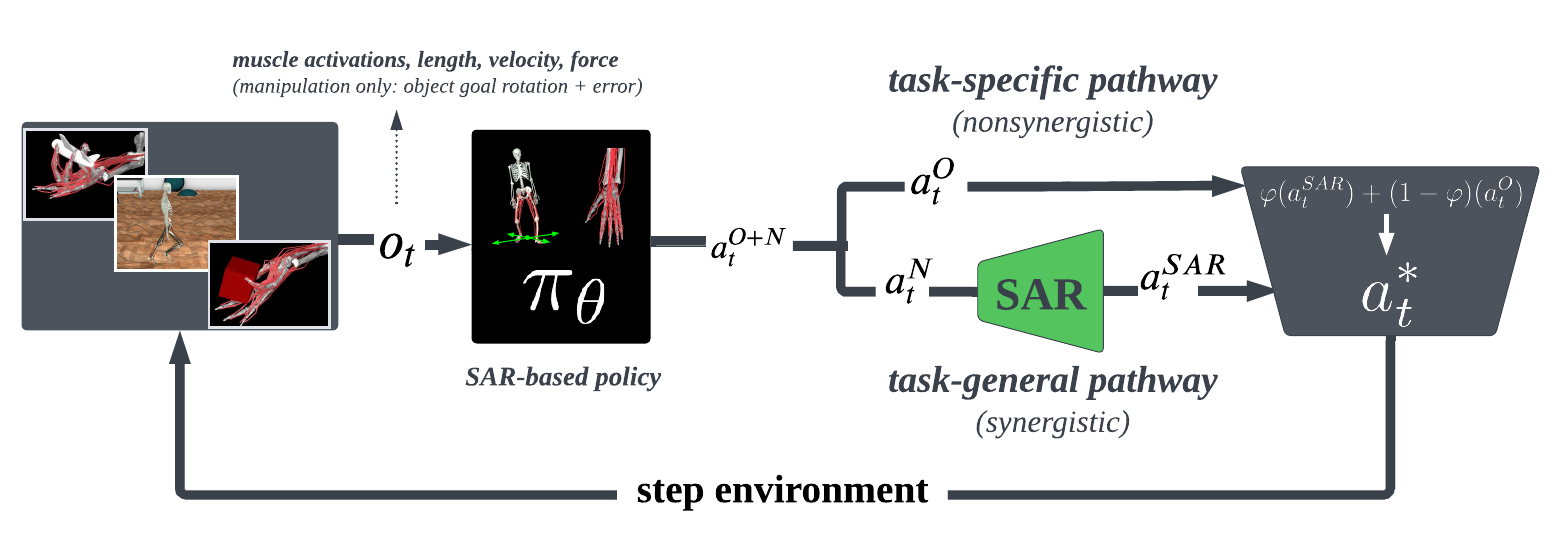}
\caption{\textbf{Policy architecture for training with SAR.} The policy takes as input $o_t$ and outputs an $O+N$-dimensional action vector, $a_{t}^{O+N}$, where $O$ is the dimensionality of the original action manifold and $N$ is the dimensionality of the synergistic manifold. The first $N$ synergistic actions are passed through $SAR(a_{t}^{N})=a_{t}^{SAR}$ to recover the synergistic muscle activations. The subsequent $O$ task-specific activations are mixed with the task-general activations using a linear blend $\varphi$ to recover the final action $a_t^*$ that steps the environment forward. Note that for \textit{MyoLegs}, it was found that purely relying on the synergistic, task-general pathway ($\varphi=1$) was sufficient for learning robust locomotion.}
\label{fig:policy_architecture}
\end{figure*}

In order to balance contributions from task-general and task-specific muscle activations, we introduce \name{} into our behavioral acquisition pipeline as an alternative weighted action representation pathway for our policy to utilize (Fig. \ref{fig:policy_architecture}). At each timestep, the policy is trained to output a $O+N$-dimensional action vector continuous in the range $[-1, 1]$, where $O$ is the dimensionality of the original action space, and $N$ is the number of synergistic modules in SAR. The first $N$ components of the action vector are input into SAR, which transforms this representation back into synergy-exploiting activations in the original muscle activation space ($a_{t}^{SAR}$). The subsequent $O$ components of the action vector ($a_{t}^{O}$) are additively combined with the synergy-exploiting activations, returning a final action ($a_{t}^{*}$) that steps the environment forward. This simple elementwise vector addition is weighted by $\varphi$, a parameter that calibrates the relative influence of the synergistic and nonsynergistic muscle activation vectors such that $a_{t}^* = \varphi(a_{t}^{SAR}) + (1-\varphi)(a_{t}^{O})$ (Fig. \ref{fig:policy_architecture}). 

Contrary to conventional wisdom that suggests a reduction the dimensionality of the action space will best aid exploration challenges, the proposed policy architecture somewhat counterintuitively increases the dimensionality of the search space. Aligned with \cite{ude_task-specific_2010, yang_dmps-based_2018, wang_generalization_2011}, our policy architecture can be viewed as providing the learning agent the option to mix behavior from task-general synergistic representational pathways (encoded by the first $N$ components of $a_t$) while simultaneously allowing for the task-specific modulations (encoded by the subsequent $O$ components of $a_t$). As is later demonstrated in our \textit{MyoHand} experiments (Sec. \ref{sec:ZeroShot}), in accordance with this intuition, our agents learn to leverage the synergistic pathways when helpful to accelerate search and proceed to rely on task-specific pathways for task customization, exploiting the benefits of both learning pathways (see Fig. \ref{fig:policy_architecture}). It is also critical to note that this architecture trivially allows for exclusively task-specific ($O$-dimensional) or task-general ($N$-dimensional) learning by setting $\varphi=0$ or $1$, respectively. Accordingly, we later find that purely relying on the synergistic, task-general pathway (i.e., $\varphi=0$) was sufficient for learning robust locomotion. As such, this policy architecture demonstrates a flexibility that renders it an attractive mechanism for using \name{} for behavior acquisition. 

\section{Acquiring Agility and Dexterity with SAR: Experimental Design}

\begin{figure}
\centering
\includegraphics[scale=.37]{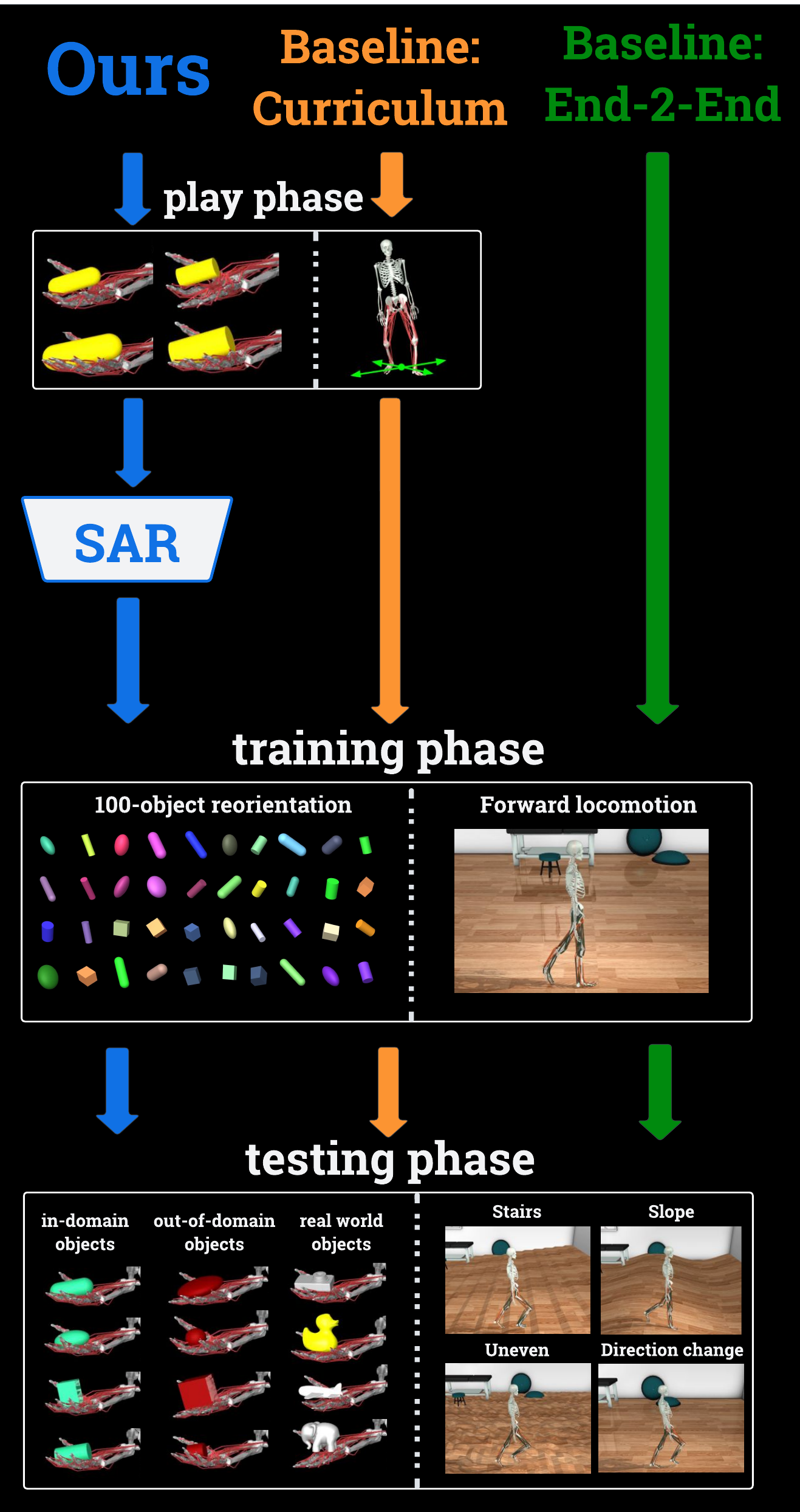}
\caption{\textbf{Experimental paradigm}. Ours (left) trains policies using a computed \name{}. Curriculum (center), leverages the play policy and extends training without using \name{}. End-2-End (right) simply trains directly on the training phase for the entire training budget. Policies are then zero- and few-shot tested on a set of out-of-domain tasks. All policies use identical hyperparameters, including total number of training samples.}
\label{Fig:trainng_paradigm}
\end{figure}

We structure our main experiment to assess the effectiveness of \name{} and our proposed policy architecture (Fig. \ref{fig:policy_architecture}) in yielding generalizable agility and dexterity. For both of these experiments using \textit{MyoLegs} and \textit{MyoHand}, we structure our experimental goals into three discrete phrases:

\begin{enumerate}
    \item \textbf{Play phase}: learn play behaviors in an environment similar to the target task in order to acquire activation data for computing \name{}.
    \item \textbf{Training phase}: acquire the target behavior(s) using the \name{}-based policy architecture (Fig. \ref{fig:policy_architecture}) in comparison to baseline learning methods (Fig. \ref{Fig:trainng_paradigm}) that do not exploit synergies. 
    \item \textbf{Testing phase}: Utilize both \name{} and the \name{}-based policy to yield zero- or few-shot generalizations to new tasks with out-of-domain environmental properties.
\end{enumerate}

Before presenting our results in Sec. \ref{sec:results}, next we outline details of our locomotion and manipulation task suites for \textit{MyoLegs} and \textit{MyoHand}, respectively, including technical task specifications as well as the play phase, training phase, and testing phase for both musculoskeletal modalities.

(a) the in-hand reorientation task suite in Sec. \ref{sec:task_specification}), (b) \name{} representation acquisition in Sec. \ref{sec:representation_acquisition}, and (c) our \name{} agent and choice of baselines in Sec. \ref{Sec:baselines}.

\subsection{Task specifications}\label{sec:task_specification}
We describe the overarching goal of the \textit{MyoLegs} and \textit{MyoHand} physiological control experiments before detailing the action, state, and reward functions used for these experiments.

\textbf{{MyoLegs control experiment}}. We task our \textit{MyoLegs} agent with learning robust locomotion using a limited training budget, beginning with a simpler flat terrain and significantly complexifying the terrain conditions later on. This task requires the continuous coordination of 80 muscles across two limbs, a highly challenging action space to learn in a small number of samples.

\textbf{{MyoHand control experiment}}. We task our \textit{MyoHand} agent with reorienting an object to a particular goal orientation. Both the specific object and the desired reorientation angle are randomly initialized at the beginning of each training episode. This task requires an agent capable of learning a highly flexible motor control policy that is robust to high variance both in object identity and desired reorientation. 

\textbf{{Musculoskeletal action space}}. For both \textit{MyoLegs} and \textit{MyoHand}, the action vector $a_t$ consists of continuous values $[-1, 1]$ for contracting each of the muscles in the musculoskeletal model (80 muscles in \textit{MyoLegs}, 39 muscles in \textit{MyoHand}. Given that muscle activations cannot be negative \cite{rabbi_non-negative_2020}, MyoSuite transforms this action vector into the range $[0, 1]$ using the sigmoid transform,$\frac{1}{1+e^{-5*(a_t - 0.5)}}$.

\textbf{Musculoskeletal state space}
For both \textit{MyoLegs} and \textit{MyoHand}, the state vector $s_{t} = \{\phi_{t}^{act}, \phi_{t}^{len}, \dot{\phi}_{t}, \phi_{t}^{F}\}$ consists of continuous values for muscle activations $\phi^{act}$, muscle length $\phi^{len}$, muscle velocity $\dot{\phi}$, and muscle force $\phi^F$. Given the object manipulations required in the \textit{MyoHand} experiment, object rotation $\psi$, object velocity $\dot{\psi}$, object goal rotation ${\psi^*}$, and object rotational error ${\psi^e}$ were also specifically included in $s_t$ for this experiment only. Note that muscle activations, force, length, and velocity were included in $s_t$ in lieu of joint positions because proprioceptive signals are instantiated in the human nervous system \cite{tuthill_proprioception_2018}, while there is no corresponding evidence that suggests precise joint positions are encoded neurophysiologically. 

\textbf{MyoHand reward function}
We employ the following reward function for learning multiobject reorientation with \textit{MyoHand}:
\begin{multline}
R(x_t,\hat{x}_t) := -\lambda_1 \| x_t^{(p)} - \hat{x}_t^{(p)} \|_2 +\lambda_2|\angle x_t^{(o)} - \hat{x}_t^{(o)}| \\
-\lambda_3 \{dropped\} -\lambda_4 \left \| \overline{m}_t  \right \|_2 + \lambda_5\{b1\} + \lambda_6\{b2\}
\label{Eq:Reward}
\end{multline}

where $\angle$ is the quaternion angle between the two current and goal orientations, $x_{t}^{(p)}$ is the object goal position, $x_{t}^{(o)}$ is the object goal orientation, $1\{dropped\}$ penalizes object dropping, $\{b1\}$ and $\{b2\}$ incentivize simultaneous rotational and positional alignment above a threshold, and $\bar{m}_{t}$ encodes overall muscle effort. 

\textbf{MyoLegs reward function}
We employ a different reward function for learning locomotion with \textit{MyoLegs}:

$$ R(t) := \lambda_1 v_t - \lambda_2 c_t + \lambda_3 r_t + \lambda_4 j_t - \lambda_5 d_t $$

where $v_t$ denotes the velocity reward (quantifying the desired velocity of the leg), $c_t$ represents the cyclic reward for the hip (aiming to encourage rhythmic movements), $r_t$ is the reward based on the reference rotation (encourages the leg to follow a given rotation pattern), $j_t$ is the joint angle reward (penalizes deviation of specific joints from their desired angles), and $a_t$ is the action magnitude, computed as the norm of the action divided by the total number of actuators (aiming to encourage efficient use of the actuators).

where $\angle$ is the quaternion angle between the two current and goal orientations, $x_{t}^{(p)}$ is the object goal position, $x_{t}^{(o)}$ is the object goal orientation, $1\{dropped\}$ penalizes object dropping, $1\{bonus1\}$ and $1\{bonus2\}$ incentivize simultaneous rotational and positional alignment above a threshold, and $\bar{m}_{t}$ encodes overall muscle effort. 

We considered a trial successful when object rotational error ${\psi^e} = |\cos{(\psi,\psi^*)}|$ was $\leq 5\%$. All trials were 50 timesteps.

\subsection{MyoHand object datasets}\label{Sec:data}
For the play, training, and testing phases of the \textit{MyoHand} reorientation experiment, objects were sampled from two datasets. The \datasetname{} consisted of four core geometries in the MuJoCo physics simulator \cite{todorov_mujoco_2012}: ellipsoid, cuboid, cylinder, and capsule (i.e., spherocylinder). Those objects could be parameterized to be scaled across the X, Y, and Z axes. Geometrical dimensions and target poses were determined given both the physiological constraints of the hand and the parameters used in related works involving manipulation \cite{caggiano_myosuite_2022}.

Additionally, a set of real-world objects was obtained by using 8 objects from the ContactDB dataset \cite{brahmbhatt_contactdb_2019}. 

\subsection{Play phases}\label{sec:representation_acquisition}
For both experiments, the purpose of the play phase is to train a policy on a simple task similar to the target task in order to extract muscle activation data that can be used to compute a \fullname{} that can be used to learn the target task (training phase).

\textbf{MyoLegs play phase}. For the \textit{MyoLegs} locomotion experiment, the play period simply consisted of training a policy on a flat walking environment for a 1.5M steps. Though this policy would not be expected to achieve locomotion in such a small number of samples, it was hypothesized that the muscle activation data yielded by this policy would be sufficient for computing a \fullname{} that would \textit{itself} ultimately facilitate locomotion.

\textbf{MyoHand play phase}. For the \textit{MyoHand} manipulation experiment, the play period consisted of training a policy to reorient a large and small variant of each of the four core MuJoCo geometries (see \ref{Sec:data}), totaling eight geometries (Table \ref{table:params}) for 1M steps. The agent is exposed randomly to all eight of the objects to learn to reorient them. It was hypothesized that this simpler environment would introduce the agent to the multiobject manipulation control problem without subjecting it to the full complexity of learning to reorient 100 discrete objects.

\subsection{Training phases}\label{sec:training_phase}
After the \textit{MyoLegs} and \textit{MyoHand} play phases, we are able to extract muscle activations from the resulting play policies, which will be transformed into a \fullname{} (see Sec. \ref{syn_ac}) and subsequently used to parameterize a new policy (Fig. \ref{fig:policy_architecture}).

\textbf{MyoLegs training phase}. We test whether \name{} acquired from a play policy 1.5M steps on a forward locomotion task is sufficient for actually yielding forward locomotion. In other words, (unlike the \textit{MyoHand} experiment) the \textit{MyoLegs} training environment is the same as that used in the play phase. However, it is critical to emphasize, as is later described in Sec. \ref{sec:results}, that the \textit{MyoLegs} play phase policy does not itself learn any meaningful walking behavior. This new policy is trained for an additional 1.5M steps.    

\textbf{MyoHand training phase}. After extracting \name{} from the play phase policy trained to reorient eight objects, we use this representation to facilitate reorientation of a much larger set of objects: we generate 25 different geometries (Table \ref{table:params}) of the 4 objects in the \datasetname{}. This set of 100 objects was randomly presented and the policy was trained for 2M steps.

\subsection{Testing phases}
After the \textit{MyoLegs} and \textit{MyoHand} training phases, we can assess the generalizability both of \name{} and of the \name{}-based policy. First, the degree to which the policy zero-shot generalizes to out-of-domain conditions can be compared to baseline approaches in order to assess the generality of what has been learned over the previous two phases. Second, we can assess how effectively the original \name{} transfers when used to few-shot train new policies on these out-of-domain environments.

\textbf{MyoLegs test phase}. We introduce four new locomotion environments in order to rigorously assess the generalizability of the locomotion-based \name{} and the associated policy: a ‘hilly’ terrain, a stair-like terrain, an uneven terrain, and a directionality change environment (i.e., a diagonal walk). We begin by zero-shot testing the trained policy on the uneven terrain to assess the generalizability of the \name{}-based policy, and we proceed use the locomotion-based \name{} to train new policies on these unseen environments to assess the generality of the original representation extracted during the \textit{MyoLegs} play phase.

\textbf{MyoHand test phase}. After training the \textit{MyoHand} on the 100-object reorientation task, the policy was zero-shot tested using three new object sets (see Fig. \ref{fig:zero_shot}). First, in an in-domain test, the policy was tested in a zero-shot inference on 1000 novel objects generated from the same parameters used for training (250 geometries for 4 objects (Table \ref{table:params}). Second, in an out-of-domain test, an additional 1000 objects (250 geometries for 4 objects) were obtained from a range of parameters outside of those used for training (Table \ref{table:params}). Finally, the generalization was tested on a \textit{RealWorldObjs} set (see Sec. \ref{Sec:data}; Fig. \ref{fig:zero_shot}).

\subsection{Baselines}\label{Sec:baselines}

For both the \textit{MyoLegs} and \textit{MyoHand} experiments, the performance of the trained \name{}-based policy (\textbf{SAR-RL}) was compared against two baselines. 
\subsubsection{Curriculum learning: RL+Curr}
First, we compared \name{} to standard curriculum learning, here named \textbf{RL+Curr}, by directly finetuning (i.e., resuming training for) the policy trained on the play phase for the main training phase. This baseline (see ‘Baseline: Curriculum,’ Fig. \ref{Fig:trainng_paradigm}) controls for whether utilizing \emph{any} form of pretraining on a simpler environment modulates performance on the more complex target task or whether the \fullname{} \emph{per se} leads to enhanced policies.

\subsubsection{End-to-end RL: RL-E2E}
In addition, we train a policy end-to-end (\textbf{End2End}, or \textbf{RL-E2E} for short, see ‘Baseline: End-2-End,’ Fig. \ref{Fig:trainng_paradigm}) on training phase task without any play phase or \fullname{}. This baseline controls for whether SAR-RL’s indirect access to additional training experience from the play phase is what is driving its performance. Accordingly, note that the total training samples and algorithm hyperparameters are identical across all three conditions.

\section{MyoLegs experimental results}\label{sec:leg_results}

We present results from the \textit{MyoLegs} experiments described in Sec. \ref{sec:task_specification}, where a \fullname{} extracted from a policy trained for a small number of steps on a flat walking task successfully enables locomotion when used to retrain on this same task. We also show that this same \name{} can be leveraged to successfully learn locomotion on a diverse set of new terrains with SOTA sample efficiency.

\subsection{\name{} enables highly sample efficient locomotion}\label{sec:loco_results}
\begin{figure}
\centering
\includegraphics[scale=.54]{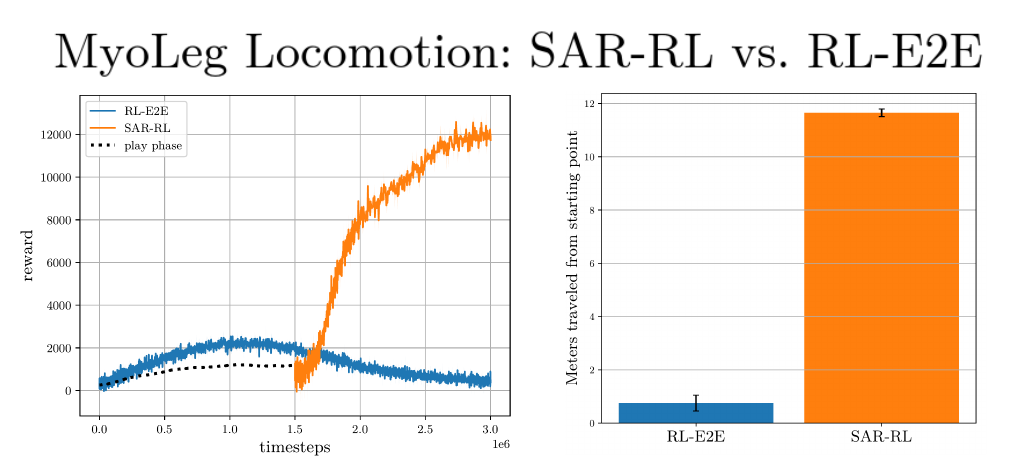}
\caption{\textbf{SAR-RL yields highly sample efficient locomotion}. Left: after 1.5M play phase steps (dotted black line), \name{} is computed and a new \name{}-exploiting policy is trained on the forward locomotion task for an additional 1.5M steps (blue). This is compared to RL-E2E (i.e., training on the forward locomotion task for all 3M steps; orange). Right: average distance traveled by the trained policies, in meters.} 
\label{Fig:loco_train}
\end{figure}

In spite of the general failure of the play phase \textit{MyoLegs} policy to yield meaningful behavior, we hypothesized that this policy may exhibit sufficient visitation of the relevant submanifolds of the action space for \name{} to contain representations useful for learning forward locomotion. Figure \ref{Fig:loco_train} demonstrates that while RL-E2E fails to yield locomotion, training with SAR-RL efficiently yields robust walking behavior. SAR-RL learns to walk $15x$ times farther (11.7m vs. 0.8m) than RL-E2E (Fig. \ref{Fig:loco_train}). 

\subsection{\name{}-RL zero-shot generalization to uneven terrain}\label{sec:loco_zeroshot}
\begin{figure}
\centering
\includegraphics[scale=.53]{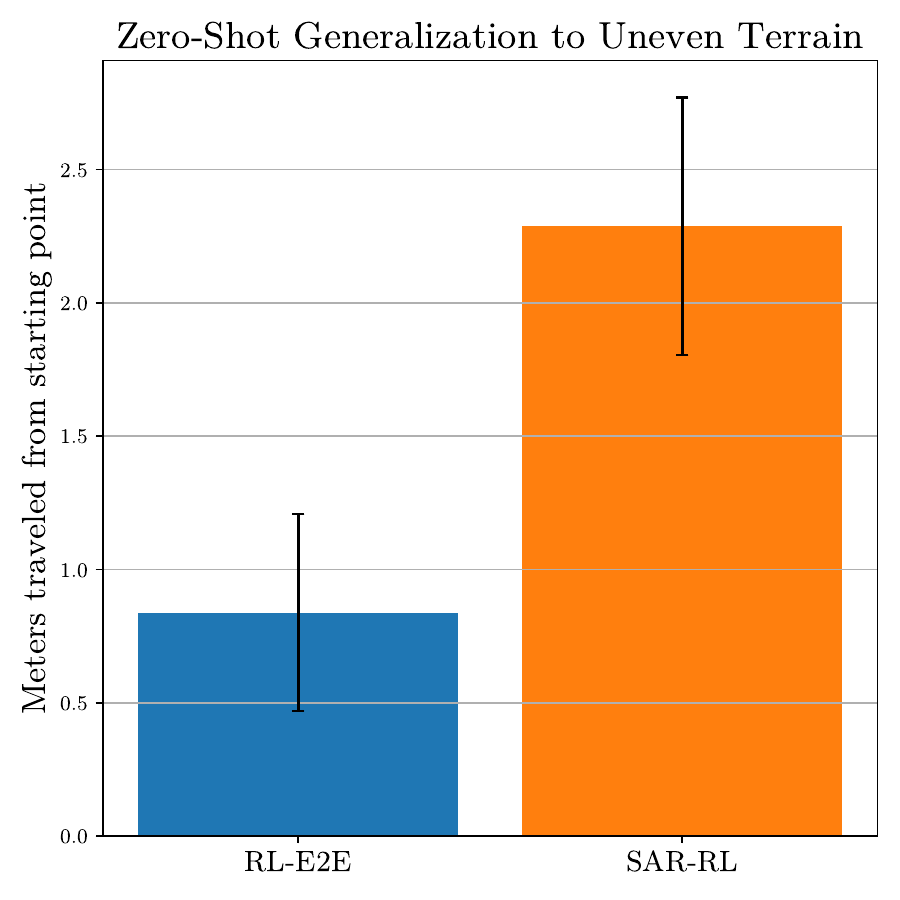}
\caption{\textbf{SAR-RL zero-shot generalization to uneven terrain}. The \name{}-based \textit{MyoLegs} training phase policy is able to walk an average of 2.3 meters zero-shot on uneven terrain, while RL-E2E is able to walk an average of 0.8 meters.} 
\label{Fig:loco_zero_shot}
\end{figure}

We test the hypothesis that \textit{MyoLegs} policies trained within the synergistic manifold (see Sec. \ref{sec:training_phase}) will demonstrate superior zero-shot generalization in unseen terrains than policies trained without \name{}. We specifically assess zero-shot generalization on uneven terrain, as is standard in the literature \cite{haarnoja_soft_2018}. We find that policies trained with SAR-RL are able to zero-shot generalize to this uneven terrain, traveling more than twice the distance as compared to the zero-shot performance of the RL-E2E baseline policy (2.3 m vs. 0.8 m; see Fig. \ref{Fig:loco_zero_shot}).

\subsection{\name{} transfers to diverse terrains}\label{sec:loco_gen}
\begin{figure}
\centering
\includegraphics[scale=.64]{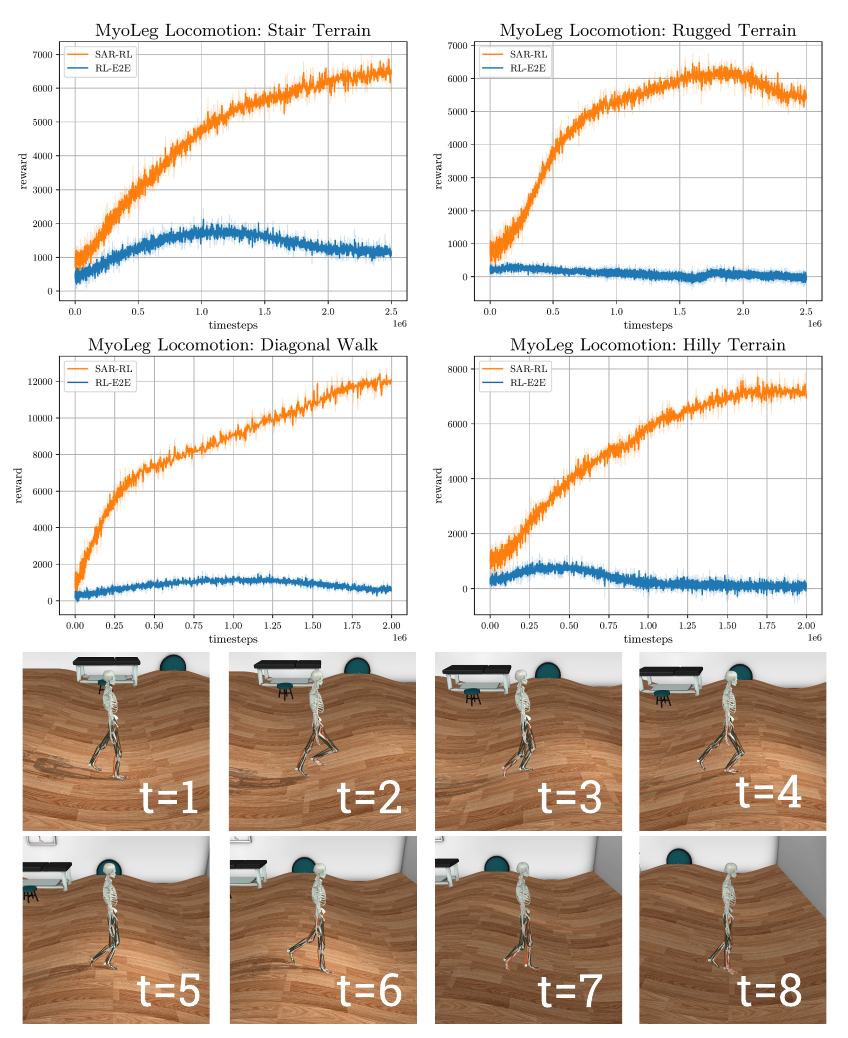}
\caption{\textbf{SAR-RL transfer learning to diverse terrains}. Across four out-of-domain environmental conditions—hilly terrain, stair-like terrain, uneven terrain, and diagonal walk—the same \name{} enables highly sample efficient locomotion, while RL-E2E fails to learn any meaningful behavior. Example frames are provided from the SAR-RL hill locomotion policy.} 
\label{Fig:legs_terrains}
\end{figure}

\begin{figure}
\centering
\includegraphics[scale=.5]{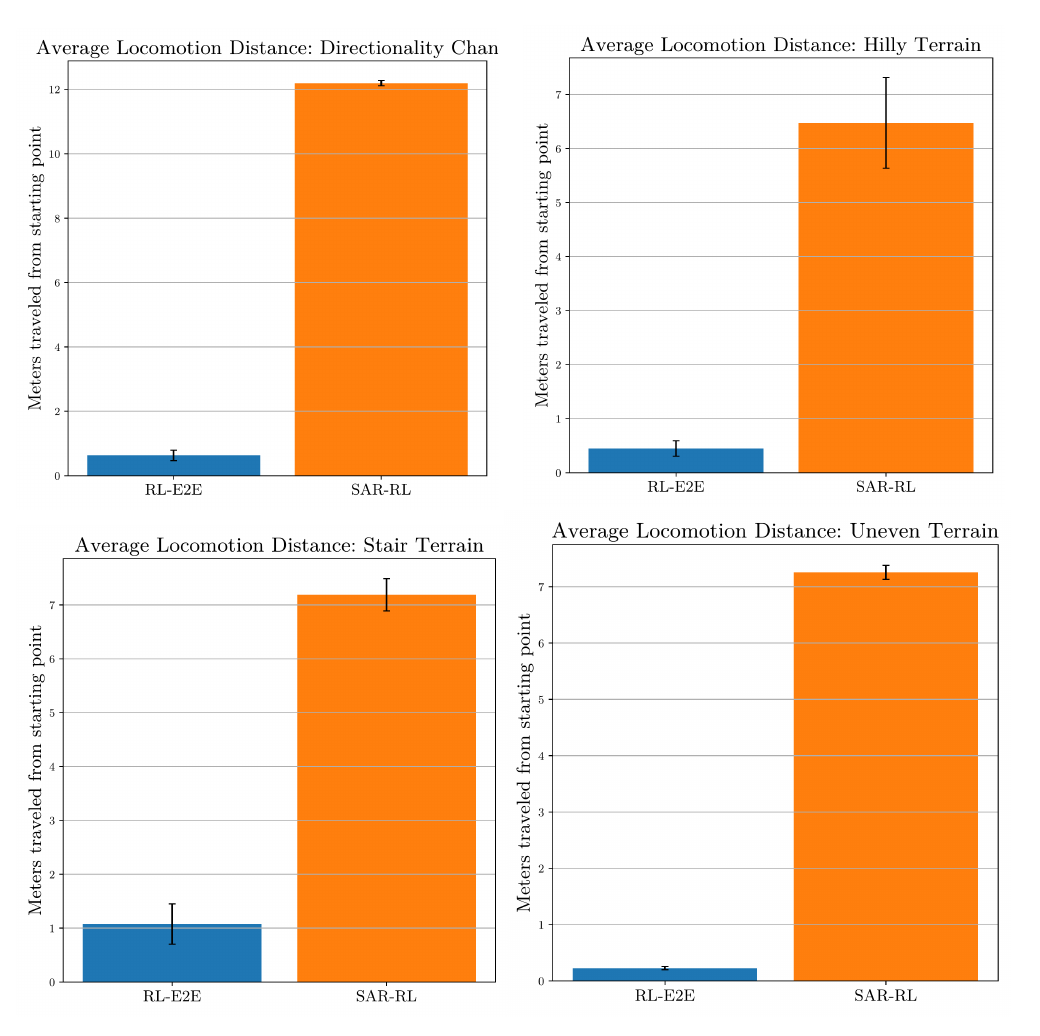}
\caption{\textbf{SAR-RL travels 16x farther than RL-E2E on average}. Across the four out-of-domain environmental conditions,RL-E2E was found to walk an average of approximately 0.5 meters, while SAR-RL was found to walk an average of approximately 8 meters, a 16-fold improvement.} 
\label{Fig:avg_dist_all}
\end{figure}

Though SAR-RL was able to relatively successfully zero-shot generalize to uneven terrain, we hypothesize that tasks with significantly out-of-distribution features such as uneven terrain, stair-climbing, hilly terrain, and directionally varied environments likely require transfer learning in order to yield robustly successful locomotion. Accordingly, we test the extent to which the same \name{} from the \textit{MyoLegs} play phase (see Sec. \ref{sec:representation_acquisition}) enables locomotion on terrains significantly different from that used to train the initial play phase policy from which \name{} was extracted. We find that this same \name{} indeed enables locomotion across diverse terrains and walking directions using only 2.5M additional samples (see Fig. \ref{Fig:legs_terrains}), while training using baseline approaches again fails to yield meaningful behavior: across the four out-of-domain environmental conditions, RL-E2E was found to walk an average of only 0.5 meters, while SAR-RL was found to walk an average of approximately 8 meters, a 16-fold improvement in performance (Fig. \ref{Fig:avg_dist_all}). 

\section{MyoHand experimental results}\label{sec:results}

We present results to show how \name{} uniquely enables learning a single policy that can solve a large set of in-hand reorientations using \textit{MyoHand} (Sec. \ref{sec:MultiTask100}). The same policy is then able to generalize to reorient objects with different shapes that were not part of the training set (Sec. \ref{sec:ZeroShot}). Also, the same synergistic representation can be used to learn in-hand reorientation on real objects (Sec. \ref{sec:Transfer}). 

\subsection{\name{} enables multiobject manipulation}\label{sec:MultiTask100}
\begin{figure}
    \centering
  \includegraphics[scale=.45]{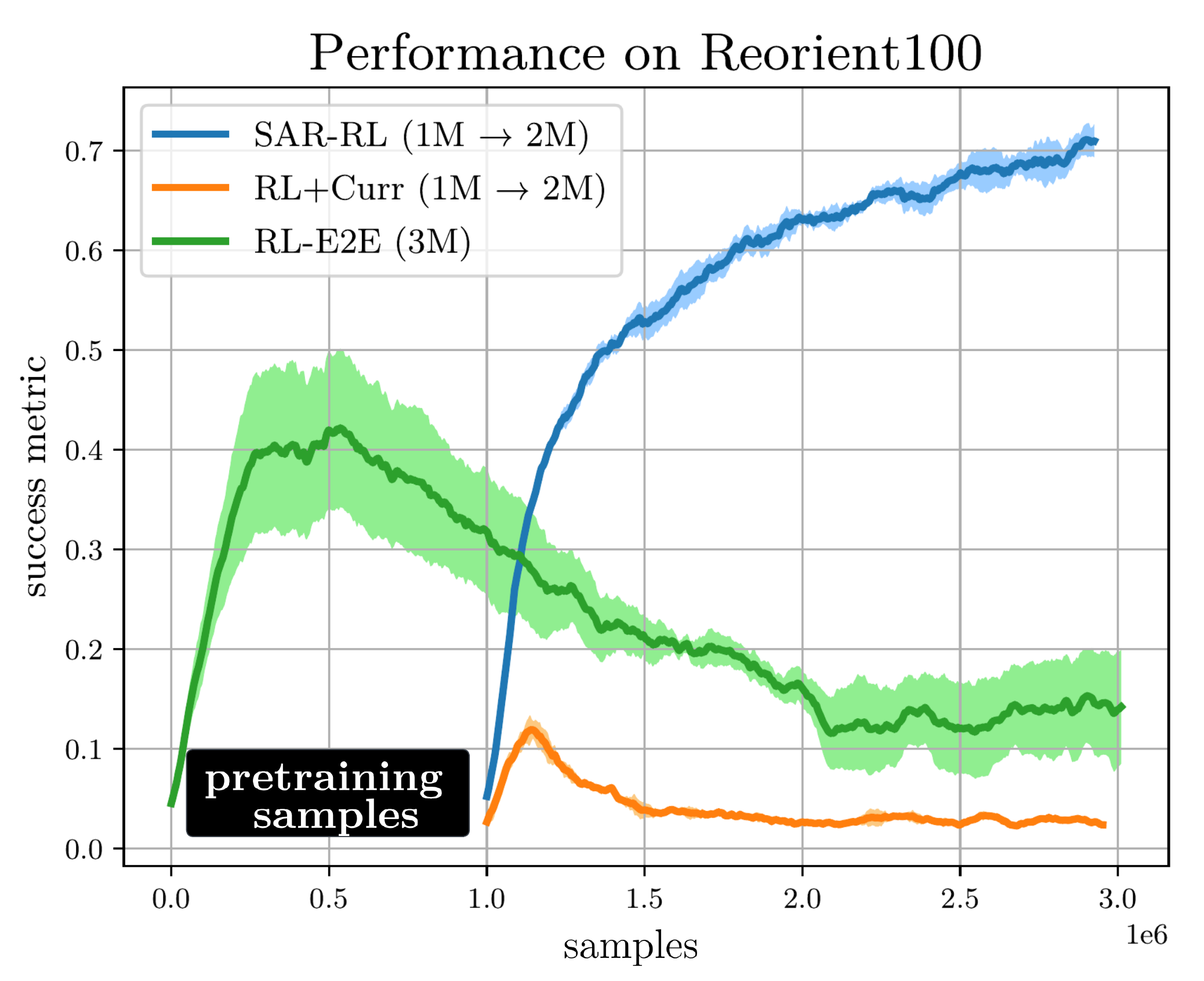}
 \caption{\textbf{Manipulation of 100 objects with \name{}}. Performance learning a multi-task policy to reorient 100 different objects by means of 3 different methods. SAR-RL, depicted in blue, uses the acquired synergistic action representation from a simpler task (1M steps) to facilitate and accelerate learning (2M steps). RL+Curr, depicted in orange, represents the use of \textit{curriculum learning} for 1M steps on a reduced number of objects, followed by finetuning this policy on the full set of 100 objects (2M steps). RL+E2E, depicted in green, represents the use of an \textit{End-2-End} policy that attempts to solve the 100 objects for the whole duration of the learning budget considered (i.e., 3M steps). SAR-RL and RL+Curr are depicted to begin at 1M timesteps in order to account for the 1M steps of pretraining embedded in both methods. The shaded area represents the variability of the policies over 3 different seeds.}
\label{fig:reorient100_main}
\end{figure}

We started with the hypothesis that learning by means of a \fullname{} (\name{}) enables learning a single policy  capable of solving a large variety of manipulation tasks. We used \name{} (extracted as described in Section \ref{sec:representation_acquisition}) to train a policy to reorient 100 different objects from the \datasetname{} (see Sec. \ref{Sec:data}).
Figure \ref{fig:reorient100_main} shows that SAR-RL is able to solve faster and achieve a significantly higher success rate ($>70\%$) than other methods. In reality, neither learning to solve directly all those tasks (RL-E2E, Fig. \ref{fig:reorient100_main}) nor pre-training (RL+Curr, Fig. \ref{fig:reorient100_main} orange line) appears to help the generalization to a larger set of objects as indicated by a success rate $<20\%$ in Figure \ref{fig:reorient100_main}. This indicates that the success at reorienting the object was due to the adoption of \name{}.

\subsection{\name{} manipulation zero-shot generalization}\label{sec:ZeroShot}
\begin{figure}
  \includegraphics[scale=.34]{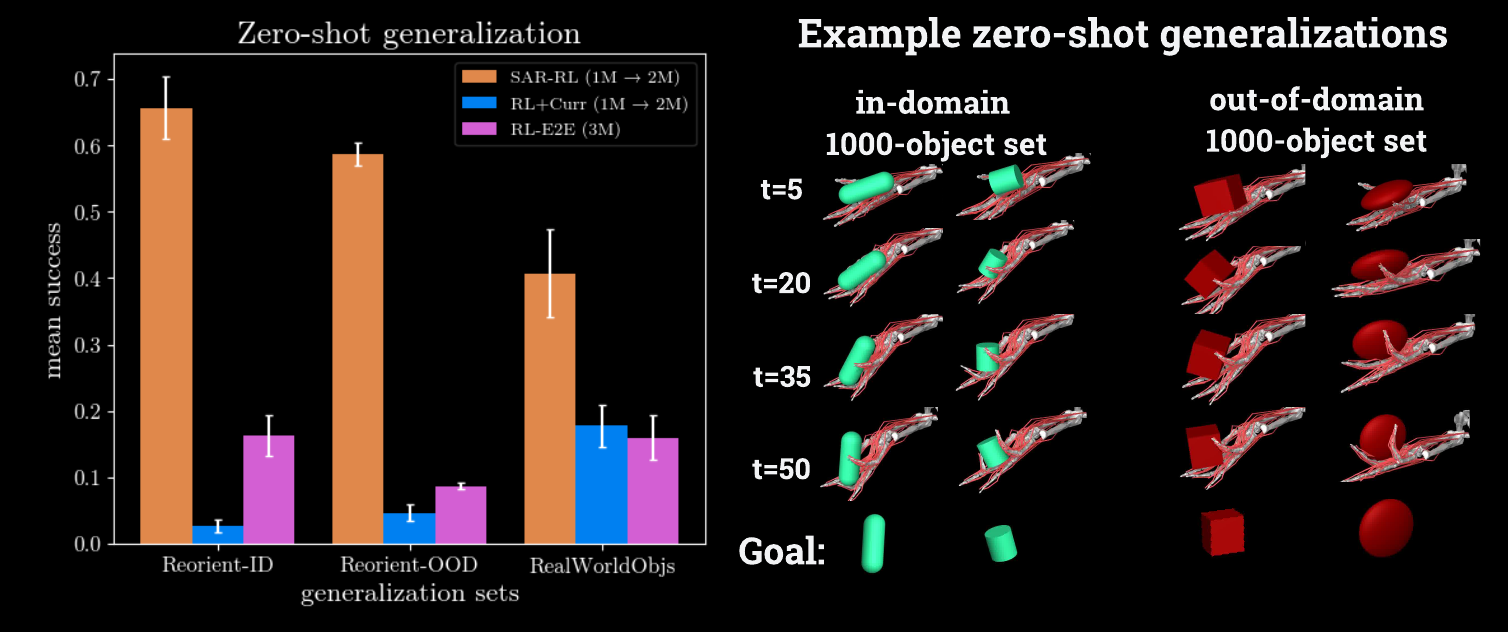}
 \caption{Left: Zero-shot testing on 1000 objects obtained within the same parameter set (in-domain) or outside the set of parameters (out-of-domain) used for generating the objects for training the \name{} based policy. In addition, a set of eight real-world objects were used. Color coded indicates the \name{}, Curriculum and End-2-End policies described in Figure \ref{fig:reorient100_main}. Right: Behavioral examples of SAR-RL's zero-shot generalization to in-domain and out-of-domain parametric objects.}
\label{fig:zero_shot}
\end{figure}

Next, we test the hypothesis that the \name{}-exploiting policy also enables superior zero-shot generalization on 3 different sets of objects (see Fig. \ref{fig:zero_shot} - left). First, from the \datasetname{} we generated a new set of 1000 in-domain objects within the same range used for the training (\textit{Reorient-ID}). Second, from the \datasetname{} we generated a new set of 1000 objects by sampling object dimensions from uniform distributions both above and below those used for the training environment (see Table \ref{table:params}). This procedure yielded objects that are both larger and smaller than—but never the same size as—those generated in the training environment (\textit{Reorient-OOD}). Third, we tested generalization on 8 different real-world objects (\textit{RealWorldObjs}). Figure \ref{fig:zero_shot} shows that the \name{}-based policy allows generalizing to reorient both in-domain and out-of-domain objects and real-world objects. Indeed, when compared to either end-to-end or curriculum learning, the generalization is $>3x$ greater for parametric objects and $>2x$ greater for the real-world objects. 

\subsection{\name{} stabilizes over learning}

\begin{figure}
\centering
\hspace{-5mm}
\includegraphics[scale=.60]{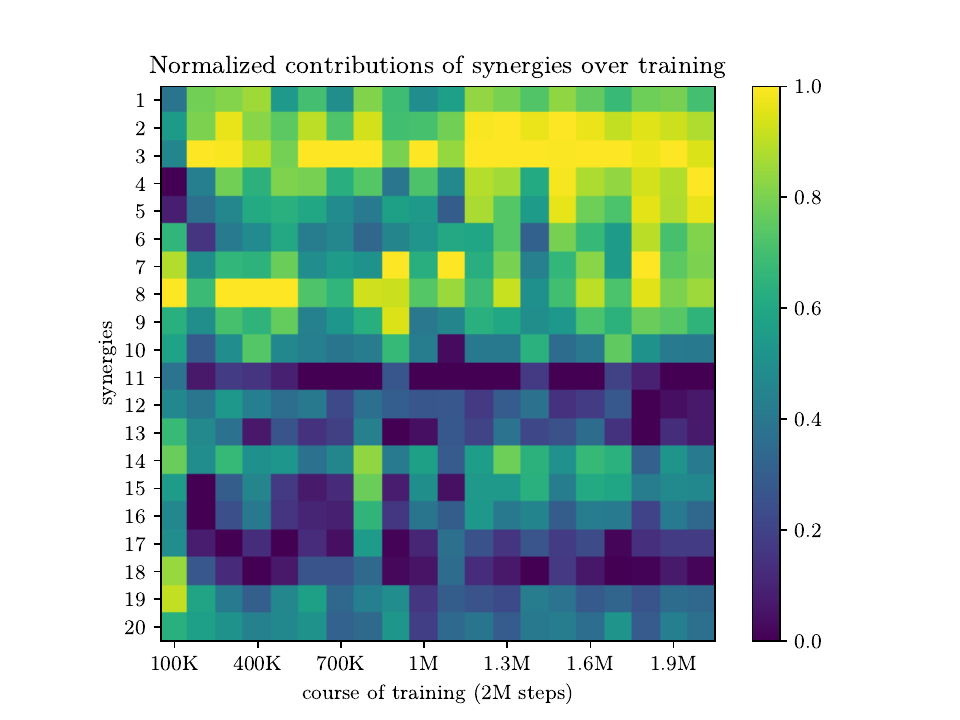}
\caption{\textbf{Relative contribution of synergies over 100 object training}. Representation of the normalized mean contributions of each synergy over 2M training steps on the 100-object reorientation environment.}

\label{Fig:norm_syn_training}
\end{figure}

Given that \name{} enables strong generalization, the question that arises is if all synergies were contributing equally to this behavior. Figure \ref{Fig:norm_syn_training} shows the normalized mean contributions of each synergy in \name{} throughout the training at different times. Throughout learning, the first 8 synergies—those that originally contributed to explaining most of the variance of the data ($~60\%$, see Fig. \ref{fig:synergies_VAF})—have a greater contribution to the final solutions as indicated by their larger weight (Fig. \ref{Fig:norm_syn_training}).

\subsection{\name{} transfer learning on Real World Objects}\label{sec:Transfer}
SAR-RL was able to generalize to both parameterized shapes and to real-world objects. Nevertheless, performance on real-world objects seems to be inferior to parameterized objects (see Fig. \ref{fig:zero_shot}, likely due to the enhanced complexity of their contact dynamics. Accordingly, we investigate if it is possible to use the \name{} to capture via few-shot learning the details of those real-world objects to reorient them. In this experiment, we leveraged only the first eight synergies as we observed to be the ones contributing most to the generalization. 

\begin{figure}
\centering
\hspace{-5mm}
\includegraphics[scale=.50]{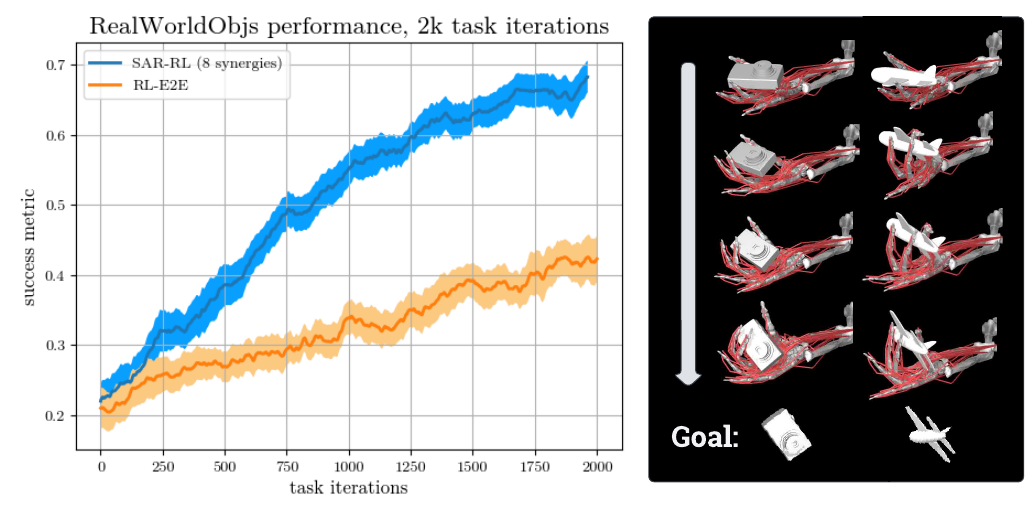}
\caption{\textbf{Few-shot learning with \name{} on RealWorldObjs}. Average policy success on the \textit{RealWorldObjs} with SAR-RL (blue) as compared to training without—i.e., RL-E2E (orange).}
\label{Fig:few_shots_fine_tuning}
\end{figure}

We find that \name{} significantly accelerates learning reorientation policies as compared to training without \name{}, approximately doubling the speed of dexterity acquisition in complex real-world object reorientation tasks (Fig. \ref{Fig:few_shots_fine_tuning}).

~~~

\section{Extending \name{} beyond physiological control}\label{sec:extendsar}
Thus far we demonstrated that SAR enables robust skill transfer for physiological control. To explore the generalizability of the SAR framework to other high-dimensional continuous control settings, we perform two additional sets of experiments on standard continuous control benchmarks. These trials successfully extend SAR to (1) the 20-DoF robotic Shadow Hand in simulation \cite{tassa_synthesis_2012}, and (2) the Humanoid-v2 gym environment \cite{1802.09464}, a 17-DoF bipedal walking task.
\subsection{\name{} enables robotic skill transfer in Shadow Hand}
First, we demonstrate SAR’s applicability to robotics problems, demonstrating that SAR enables approximately 2x faster learning of randomized object reorientation on Shadow Hand compared against end-to-end RL ({see Fig. \ref{Fig:shadowhand}}. 

\subsubsection{Dexterity transfer between geometries}
Our first Shadow Hand result demonstrates the generalizability of SAR using synergies computed from policies trained on the official gymnasium-robotics ‘HandManipulateEggRotate-v1’ environment. It is important to note that the control actions in this environment are absolute angular positions of the actuated joints rather than continuous muscle activations (as in the MyoHand). This requires that the synergistic action representation extracted from this environment would comprise coordinated joint movements rather than muscle co-contractions. 

In this experiment, we first train an end-to-end policy to manipulate an egg to a randomized target orientation. We then compute and extract SAR by rolling out this policy. Next, we use SAR (see Fig. \ref{fig:policy_architecture}) to train reorientation policies on two unseen geometries: capsules and cylinders. In comparison to end-to-end RL on cylinder and capsule reorientation, we find that SAR-RL achieves approximately 2x better performance using the same number of samples (see Fig. \ref{Fig:shadowhand}). 


\subsubsection{Replication of Reorient100 experiment}
\begin{figure}
\centering
\includegraphics[scale=.4]{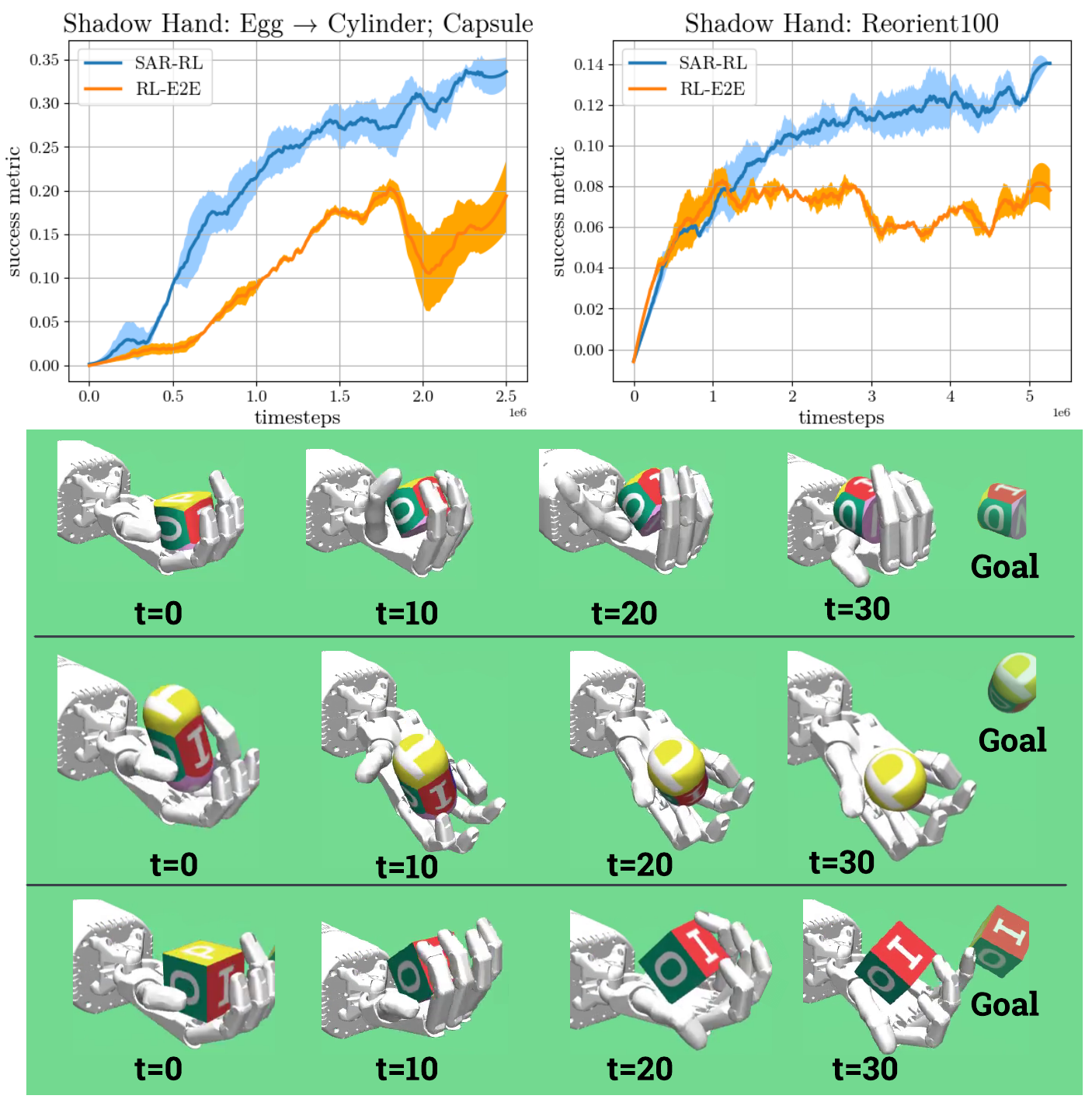}
\caption{\textbf{\name{}-enabled dexterity on Shadow Hand}. Top left: Using synergies computed from an egg manipulation policy, SAR-RL achieves approximately 2x better performance on randomized capsule and cylinder reorientation compared to end-to-end RL. Top right: SAR-RL achieves approximately 2x better performance on the Reorient100 task compared to end-to-end RL with synergies computed from an 8-object reorientation policy. Bottom: example manipulations of the policy trained with SAR-RL on Reorient100.}
\label{Fig:shadowhand}
\end{figure}

Our second Shadow Hand result reimplements the Reorient100 experiment in the robotic task setting. As in the MyoHand experiment (see Fig. \ref{Fig:trainng_paradigm}), we use an eight-object reorientation task (2 x 4 geometric shapes) to train a policy, compute SAR from rollouts of this trained policy, and use the acquired representation to facilitate learning of the significantly more challenging 100-object (25 x 4 geometric shapes) reorientation task (Fig. \ref{Fig:shadowhand}). We find that, compared to end-to-end training, SAR-RL achieves approximately 2x better performance using the same number of samples on the 100-object task.

\subsection{\name{} efficiently yields robust humanoid locomotion}
\begin{figure}
\centering
\hspace{-5mm}
\includegraphics[scale=.32]{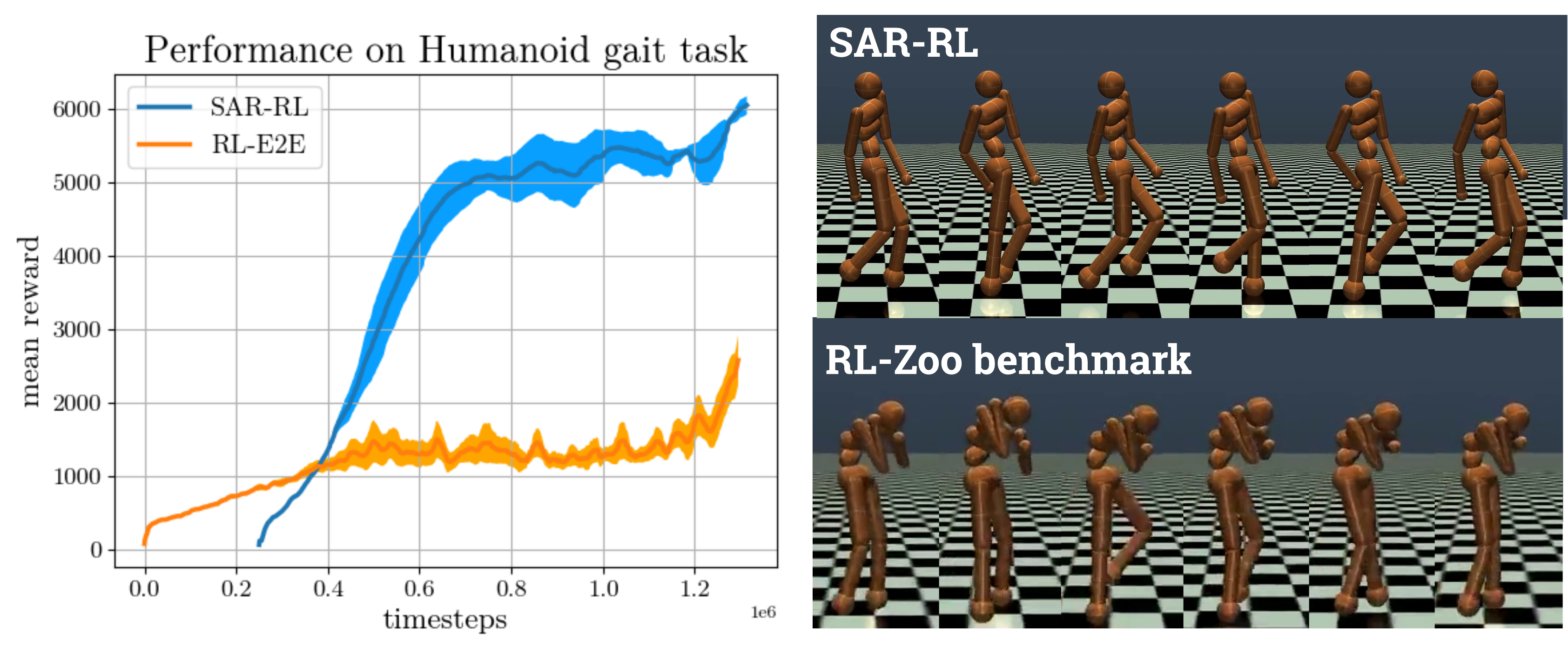}
\caption{\textbf{\name{}-enabled locomotion on Humanoid}. Left: SAR-RL yields 3x better performance on Humanoid-v2 than RL-E2E in the same number of total timesteps. Right: behavioral comparison of the locomotion behavior yielded by SAR-RL in comparison to the \href{https://huggingface.co/sb3/tqc-Humanoid-v3}{top RL-Zoo benchmark}.}
\label{Fig:humanoid}
\end{figure}

We proceed to extend SAR to full-body motor control on the Humanoid-v2 gym environment. We find that by simply training on this environment for a small number of steps, computing and extracting SAR from this early policy, and using this representation to parameterize the training of a new policy on the same environment, we are able to reach SOTA performance in only 1.25M total training steps (Fig. \ref{Fig:humanoid}), one to two orders of magnitude more efficiently than competing approaches documented in the literature (see Table \ref{table:humanoid}). Of note, we also find that training with SAR yields a significantly more natural gait compared to baseline approaches (e.g., consistent cadence and step length; presence of stance and swing phases; arm swinging and upright posture).	

\begin{table}[ht]
\centering
\begin{tabular}{>{\raggedright}p{2.25cm} >{\centering}p{2cm} >{\centering\arraybackslash}p{2cm}}
\toprule
Experiment & Algorithm & Returns at 1.25M samples \\
\midrule
SAR (ours) & SAR+SAC & \textbf{6104} \\
Wang \& Ni, 2020 & meta-SAC & 5610 \\
Peng et al, 2010 & SAC & 4100 \\
Wu et al, 2022 & SAC & 4600 \\
Toklu et al, 2020 & PGPE & 715 \\
RL-Zoo baseline & SAC & 1552 \\
AI2 baseline & PPO & 1120 \\
\bottomrule
\end{tabular}
\caption{Comparing SOTA performance on Humanoid-v2 at 1.25M samples.}
\label{table:humanoid}
\end{table}

\section{Discussion} 
Leveraging SAR computed from simpler tasks as a weighted action representation for learning more complex tasks is demonstrated to be a uniquely effective means for enabling strong performance (a) for bipedal locomotion over diverse terrains, (b) for in-hand multiobject manipulation, and (c) for high-dimensional control problems beyond the scope of physiological control. Alternative solutions that did not leverage synergies did not yield comparably effective or generalizable policies. What accounts for this striking difference in performance?

Framing muscle synergies as an embodied form of transfer learning proves useful for understanding how they enable strong performance on high-dimensional continuous control tasks. It is well-known that in humans, skill transfer learning occurs across the entire lifespan and is an important contributing factor to the uniquely sophisticated behavior of humans \cite{bassett_understanding_2011}. Such learning prevents human learners from being required to 'reinvent the wheel' with each new behavioral repertoire by leveraging already-existing stable action representations accrued during lifelong learning. Analogously, in our experiments, we show that (a) the exact same set of synergies can be reused and effectively finetuned for diverse task settings, and that (b) normalized contributions of the synergies stabilize over training to utilize a core subset of coordinated muscle co-contraction patterns (Fig. \ref{fig:reorient100_main}). Together, these findings constitute strong evidence that a physiologically coherent action representation is transferred from one task to the next. 

Additionally, the successful extension of SAR to robotic and humanoid control demonstrates that the utility of the representation is not constrained to musculoskeletal control and can more generally enable robust learning in high-dimensional systems. Taken as a whole, these investigations demonstrate that leveraging synergies—a core mechanism of human motor organization and adaptation—enables a degree of dexterity and agility that was otherwise not reached using baseline learning methods. Accordingly, we propose that learning high-dimensional control using \fullname{}\textit{s} is a promising mechanism for instantiating skill transfer learning and ultimately bootstrapping towards a generalist embodied agent.

\section{Limitations and future research} \label{limitations}
There are a number of limitations and opportunities for future research from the present study. While the \textit{MyoHand} and \textit{MyoLegs} are physiologically accurate from a structural perspective, there are still fundamental differences between the state and action spaces employed in human motor learning and those of the musculoskeletal models utilized in this research. The most obvious difference is that the these models lack two sensory modalities: touch and sight. Previous work suggests that both of these sensory pathways in the nervous system help facilitate human-level dexterity and agility \cite{johansson_19_1996} and that a loss of touch in particular can severely compromise dexterity in humans \cite{luukinen_effect_2021}. These facts reveal both a strength and a limitation of this investigation: it is limited insofar as the policies yielded from the synergistic learning paradigm still should not be expected to precisely mirror the behavioral style of human learners given dissimilar sensory data. However, this perceptual deficit also reveals a strength of the synergistic learning paradigm: namely, that sight and touch \emph{were not required} to yield successful control policies for these complex tasks.

This investigation naturally enables a number of opportunities for future research. First, the method for computing and instantiating muscle synergies developed in this investigation is entirely task-agnostic, meaning that our framework could be conceivably leveraged for any musculoskeletal task set. It is also worth noting that the strategy of using normalized ICAPCA to compute muscle synergies is potentially viable for yielding useful, dimensionality-reduced representations that can be conceivably leveraged for any RL skill transfer problem. In terms of object-level future opportunities, more formally studying the extent to which muscle synergies can be utilized as an embodied form of transfer learning remains highly underexplored and is a promising area of future research.

\footnotesize
\bibliographystyle{plainnat}
\bibliography{test.bib}

\onecolumn
\newpage
\appendix
\section{Appendix}
\counterwithin{figure}{section}
\counterwithin{table}{section}
\normalsize

\subsection{Real world objects set used in this study}
We begin by presenting the complete set of real world object we used in our experiments in Sec.\ref{sec:ZeroShot}.

\begin{figure}[h]
\centering
\includegraphics[scale=.3]{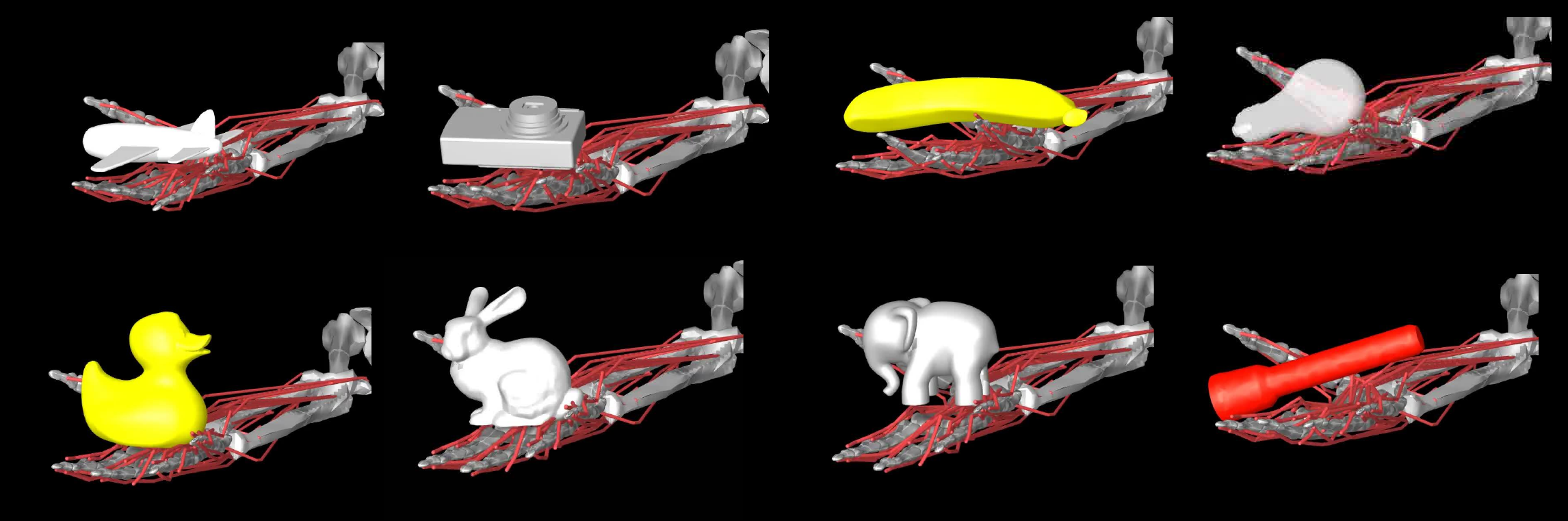}
\caption{Eight objects used for RealWorldObjs data set (see Sec. \ref{sec:ZeroShot}).}
\label{fig:realworld}
\end{figure}

\subsection{Ablations}\label{sec:Ablations}
We present a series of ablation experiments to understand how our policy and representation design choices in the Reorient100 experiment impacted results.

\subsubsection{Relative contribution of synergies to the solution}

\begin{wrapfigure}{r}{0.6\textwidth}
\centering
\vspace{-10mm}
\includegraphics[scale=.5]{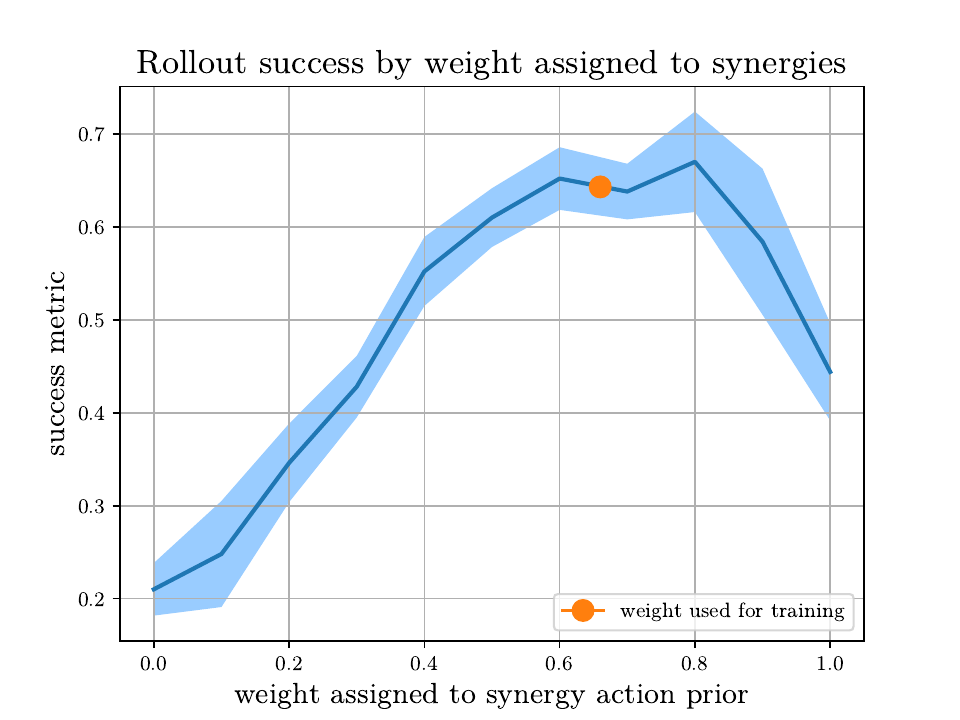}
\caption{\textbf{Post hoc modulation of weight assigned to \name{}} Analysis of the relationship of $\varphi$ to the success rate. The value of $\varphi$ used for the main experiment is depicted in orange.}
\label{Fig:weight_mod}
\vspace{-18mm}
\end{wrapfigure}

First, we investigate the choice and sensitivity of our blend weight ($\varphi$, see Sec. \ref{sec:behav_acquisition}) between the synergistic and non-synergistic dimensions. We analyzed the contribution of \name{} in terms of success rate in Figure \ref{Fig:weight_mod}. While an action contribution of \name{} between 0.6 and 0.8 does not change greatly the results, the solution greatly benefits from the use of \name{}. Indeed, choosing a very weak contribution or no contribution of the synergies negatively impacts the overall performance, leading to success rate $<0.25$. This result indicates that SAR is in fact utilized by our trained policy.

\subsubsection{Relative contribution of most vs least meaningful synergies}

\begin{wrapfigure}{r}{0.6\textwidth}
\centering
\vspace{2mm}
\includegraphics[scale=.5]{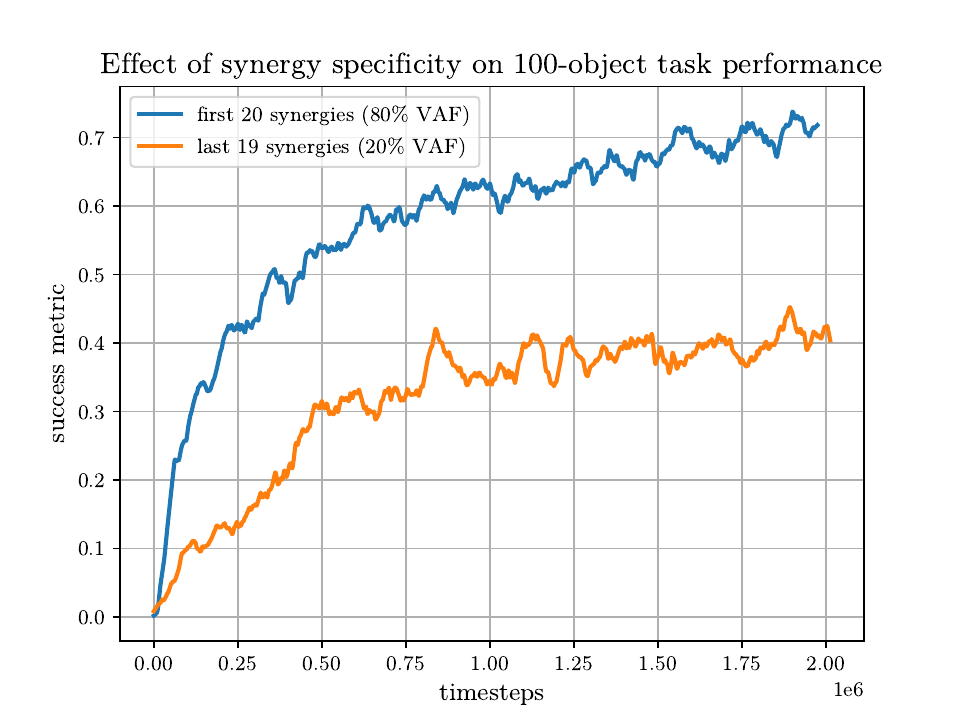}
\caption{\textbf{Comparison SAR performance using more vs. less informative synergies}. Utilizing synergies that explain more variance in a muscle activation dataset yields more robust policies. It should be noted that both of these policies significantly outperform the two baselines tested in this investigation (see Fig. \ref{fig:reorient100_main}).}
\label{fig:most_vs_least_syn}
\end{wrapfigure}

We have demonstrated (Fig. \ref{Fig:norm_syn_training}) that during training, the earlier synergies that explain most of the variance of the original signal have an outsized contribution to the final policy. Nevertheless, a question arises whether any choice of synergies produce the same results? Here, we tested the contribution of the synergies that contribute most versus the ones that contributed the least to the original tasks on which the \name{} was built (see Appendix Fig. \ref{fig:synergies_VAF}).
Figure \ref{fig:most_vs_least_syn} shows that when the least informative synergies for the pre-training task were chosen to compute \name{}, their ability to facilitate the learning of a larger set of reorientation was heavily compromised, reaching about half of the performance when more significant synergies were utilized for \name{}. Overall, this result is consonant with the observation of Figure \ref{Fig:norm_syn_training} where the use of synergies that capture a disproportionate amount of the information from the signal—i.e. first synergies have higher VAF (see Fig. \ref{fig:synergies_VAF})—also contribute a disproportionate amount to the solution.

\subsection{Dimensionality (N) of the \fullname{} (\name{})}

\begin{wrapfigure}{r}{0.6\textwidth}
    \centering
    \vspace{-7mm}
    \includegraphics[trim={0 0 0 5mm},clip, width=0.4\textwidth]{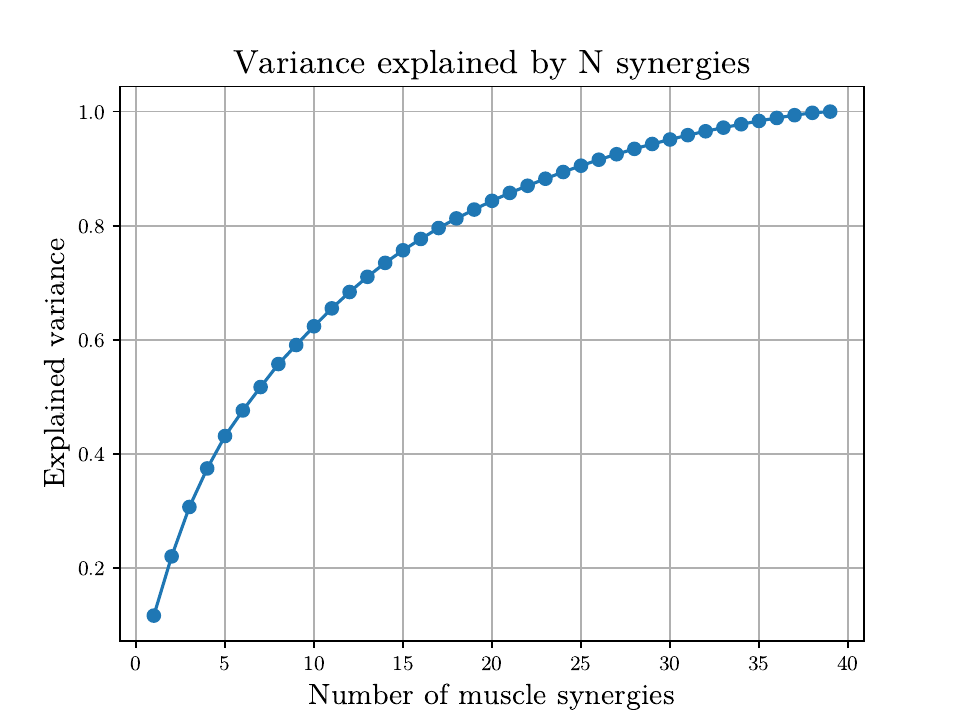}
    \caption{Proportion of variance in muscle activation data explained from play phase policy as the number of computed muscle synergies increases.}
    \label{fig:synergies_VAF}
    \vspace{-18mm}
\end{wrapfigure}

 For the purposes of this investigation, SAR is constructed from the first 20 synergies computed from the play phase described in Sec. \ref{syn_ac}. This quantity of muscle synergies was selected because it balances (a) explaining a large proportion of the variance ($>80\%$) in the initial activation data while simultaneously (b) reducing the dimensionality of the original data by a factor of 2 (see Fig. \ref{fig:synergies_VAF}). We also ablate in Fig. \ref{fig:most_vs_least_syn} that training with synergies that explain more variance in the the muscle activation data leads to improved performance as compared to training with later synergies.

\vspace{1.5cm}
\subsection{\fullname{} vs random action representation}

\begin{wrapfigure}{r}{0.6\textwidth}
    \centering
    \vspace{-5mm}
    \includegraphics[trim={0 0 0 5mm},clip, width=0.4\textwidth]{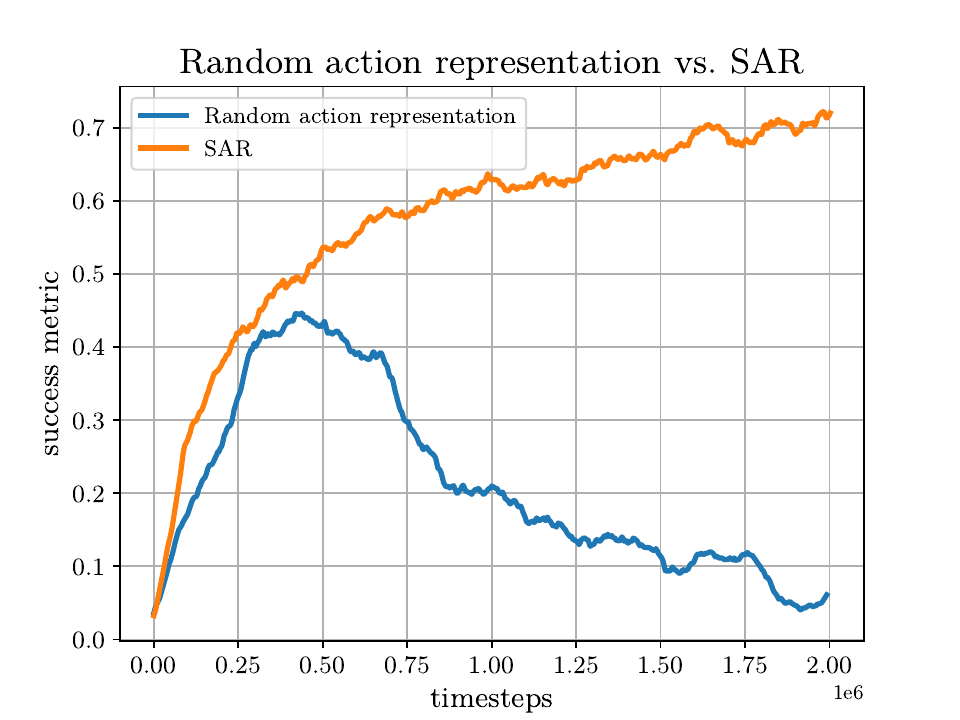}
    \caption{Training performance when SAR is computed as described in Sec. \ref{syn_ac} as compared to initializing a random representation.}
    \label{fig13}
    \vspace{-30mm}
\end{wrapfigure}

Additionally, in order to further demonstrate the utility of the latent representation captured by SAR, we compare performance between training with SAR and training with randomized PCA and ICA component matrices that share the same dimensions as SAR (see Fig. \ref{fig13}). This analysis demonstrates that randomized representations are ineffective for learning the multi-object reorientation task, which further indicates that SAR \textit{per se} has a representational utility.
\vspace{5mm}

\vspace{2cm}
\subsection{Effect of blend weight ($\varphi$) on effectiveness of \name{}}
\begin{wrapfigure}{r}{0.6\textwidth}
    \centering
    \vspace{-15mm}
    \includegraphics[trim={0 0 0 5mm},clip,width=0.4\textwidth]{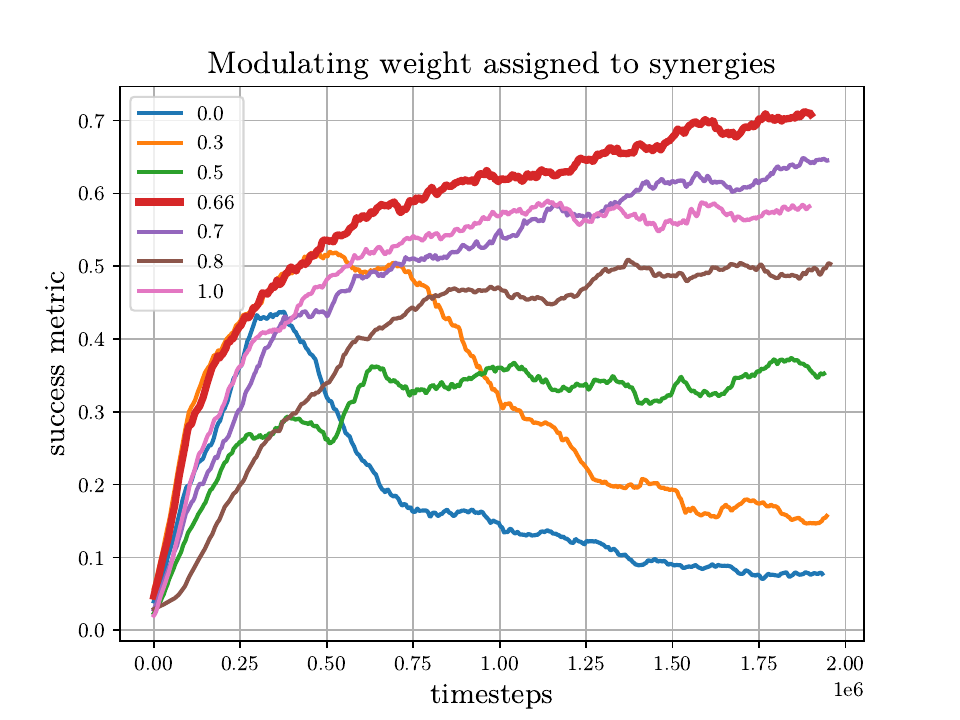}
    \caption{Training performance at different values of $\varphi$. In bolded red is $\varphi^*=.66$, the value used for the main experiment. We find that as $\varphi$ is increased from this value (i.e., SAR is given more weight in $a_t^*$), performance decrements slightly (0.7 → 0.5-0.6 success rate). As $\varphi$ is decreased from this value (i.e., SAR is given less weight in $a_t^*$), performance decrements significantly (0.7 → 0.1-0.4 success rate).}
    \label{fig:blend_wt_ablation}
\end{wrapfigure}
\normalsize An important consideration for our investigation was the optimal blend weight, $\varphi^*$, for mixing synergistic and nonsynergistic activations for each final action, $a_t^*$ (see Fig. \ref{fig:policy_architecture}). In  Fig. \ref{fig:blend_wt_ablation} we present an ablation study involving various choices of blend weight. We observe that an approximate ratio of $2/3$ SAR activations, $1/3$ task-specific activations (i.e., $\varphi \approx .66$) facilitated best performance. More specifically, we find that as as $\varphi$ is increased from this value and SAR is more heavily weighted in $a_t^*$, performance decrements slightly (0.7 success rate → 0.5-0.6 success rate). As $\varphi$ is decreased from this value and SAR is less heavily weighted in $a_t^*$, performance decrements significantly (0.7 success rate → 0.1-0.4 success rate) (see Fig. \ref{fig:blend_wt_ablation}). 

This result can be thought as \textit{proactively} searching over values of $\varphi$—in other words, different values can be specified for $\varphi$ prior to training, and variation in performance can be compared. In contrast, we can also consider \textit{retroactively} searching over values of $\varphi$—in other words, SAR-RL trained at some value (.66 in this case) of $\varphi$, but the $\varphi$ passed into the trained policy can be modulated during test to assess the impact of SAR on $a_t^*$ on the policy. This latter approach is precisely what was done for SAR-RL in Fig. \ref{Fig:weight_mod}. 

Thus, comparing proactive and retroactive modulation of $\varphi$ yields a similar result: training performance is best at $\varphi \approx .66$, decreases slightly as $\varphi > .66$, and decreases significantly as $\varphi < .66$. This result indicates that the combination of synergistic and nonsynergistic activations leads to best performance (see Sec. \ref{syn_ac}), but that more heavily weighting SAR within this mix is required for achieving this level of performance. It should be emphasized that at the extremes, $\varphi = 0$—i.e., training with no SAR—leads to poor performance (success rate $< 0.2$), while $\varphi = 1$—i.e., training only with SAR—leads to significantly improved but still suboptimal performance (success rate $\approx 0.6$) (see Fig. \ref{fig:blend_wt_ablation}).

\subsection{Parameter and hyperparameter selection}

We present parameters utilized for the training and testing regimes of our main experiments (see Secs.  \ref{sec:ZeroShot}; \ref{sec:representation_acquisition}). We selected parameters for the X, Y, and Z axes of the parametric objects used in Secs. \ref{sec:ZeroShot} and \ref{sec:representation_acquisition} by sampling over ranges for each geometry that were conducive to in-hand manipulation (i.e., excluding shapes that were too large to fit in-hand or too small to properly manipulate). (see Table \ref{table:params}; Fig. \ref{Fig:trainng_paradigm})

Additionally, we utilize hyperparameters for SAC (Table \ref{table:params_SAC}) inpsired by previous successful utilizations of this learning algorithm for continuous control in robotic RL \cite{raffin_smooth_2021}. We also include a linearly-scheduled learning rate to prevent unstable learning as the end of the training regime is approached.

\begin{table}[h!]
\centering
\caption{Parameter distributions used for generating pretraining, in-domain, and out-of-domain parametric objects (see Sec. \ref{sec:ZeroShot}). Note that for out-of-domain samples, either a larger or smaller uniform distribution was first randomly selected from which to sample parameters for each specified axis.}
\label{table:params}
\begin{tabular}{>{\raggedright}p{1.5cm} >{\raggedright}p{3cm} >{\raggedright}p{3.2cm} >{\raggedright\arraybackslash}p{3.2cm}}
\toprule
Geometry & Pretraining & In-domain & Out-of-domain\\
\midrule
Ellipsoid & obj1=[0.011,0.025,0.025],\newline obj2=[0.019,0.040,0.040] & $X\sim\mathcal{U}_{[0.008,0.02]}$,\newline $Y,Z\sim\mathcal{U}_{[0.020,0.045]}$ & $X\sim\mathcal{U}_{[0.008,0.02]}$,\newline $Y,Z\sim\mathcal{U}_{[0.015,0.020]} \lor \mathcal{U}_{[0.045,0.050]}$ \\
\addlinespace
Box & obj1=[0.017,0.017,0.017],\newline obj2=[0.023,0.023,0.023] & $X,Y,Z\sim\mathcal{U}_{[0.015,0.025]}$ & $X,Y,Z\sim\mathcal{U}_{[0.025,0.030]}\lor\newline \mathcal{U}_{[0.010,0.015]}$ \\
\addlinespace
Capsule & obj1=[0.013,0.025,0.025],\newline obj2=[0.019,0.040,0.040] & $X\sim\mathcal{U}_{[0.010,0.022]}$,\newline $Y,Z\sim\mathcal{U}_{[0.020,0.045]}$ & $X\sim\mathcal{U}_{[0.010,0.022]}$,\newline $Y,Z\sim\mathcal{U}_{[0.015,0.020]} \lor \mathcal{U}_{[0.045,0.050]}$ \\
\addlinespace
Cylinder & obj1=[0.013,0.025,0.025],\newline obj2=[0.019,0.040,0.040] & $X\sim\mathcal{U}_{[0.010,0.022]}$,\newline $Y,Z\sim\mathcal{U}_{[0.020,0.045]}$ & $X\sim\mathcal{U}_{[0.010,0.022]}$,\newline $Y,Z\sim\mathcal{U}_{[0.015,0.020]} \lor \mathcal{U}_{[0.045,0.050]}$ \\
\bottomrule
\end{tabular}
\end{table}

\begin{table}[h!]
\centering
\caption{Hyperparameters used for training with SAC across all experiments. The stable-baselines3 implementation of SAC was utilized for this investigation (see Sec. \ref{sec:problem_formualtion}).}
\label{table:params_SAC}
\begin{tabular}{|>{\raggedright}p{2.5cm}|>{\raggedright\arraybackslash}p{3cm}|}
\hline
\multicolumn{2}{|c|}{\textbf{Soft Actor-Critic (SAC) hyperparameters}} \\
\hline
Policy type & MlpPolicy \\
\hline
Actor architecture & $[400,300]$ \\
\hline
Critic architecture & $[400,300]$ \\
\hline
Learning rate & $linearschedule(.001)$ \\
\hline
Start learning & $t=3000$ \\
\hline
Batch size & 256 \\
\hline
$\tau$ & .02 \\
\hline
$\gamma$ & .98 \\
\hline
\end{tabular}
\end{table}

\end{document}